\documentclass[acmsmall,nonacm]{acmart}

\usepackage{bm}
\usepackage{multirow}
\usepackage{tabularx}
\usepackage{subfig}
\usepackage{natbib}
\usepackage{url}
\usepackage{dsfont}
\usepackage{soul}
\usepackage{xcolor}

\def\bcbaux#1#2 #3\endbcb{%
  \colorbox{#1}{\strut#2}%
  \ifx\relax#3\relax\def\next{}\else%
    \colorbox{#1}{ \strut}%
    \allowbreak%
    \def\next{\bcbaux{#1}#3\endbcb}%
  \fi%
  \next%
}
\usepackage{color}
\usepackage{tikz}
\usepackage[edges]{forest}
\definecolor{hiddendraw}{RGB}{205, 44, 36}
\definecolor{hidden-blue}{RGB}{194,232,247}
\definecolor{hidden-orange}{RGB}{243,202,120}
\definecolor{hidden-yellow}{RGB}{242,244,193}
\tikzstyle{mybox}=[
    rectangle,
    draw=hiddendraw,
    rounded corners,
    text opacity=1,
    minimum height=1.5em,
    minimum width=5em,
    inner sep=2pt,
    align=center,
    fill opacity=.5,
    ]

\AtBeginDocument{%
  \providecommand\BibTeX{{%
    \normalfont B\kern-0.5em{\scshape i\kern-0.25em b}\kern-0.8em\TeX}}}

\setcopyright{acmcopyright}
\copyrightyear{2018}
\acmYear{2018}
\acmDOI{XXXXXXX.XXXXXXX}

\acmJournal{JACM}
\acmVolume{}
\acmNumber{}
\acmArticle{}
\acmMonth{}

\begin{document}

\title{Membership Inference Attacks on Machine Learning: A Survey}
\author{Hongsheng Hu}
\affiliation{%
   \institution{The University of Auckland}
   \country{New Zealand}}
\email{hhu603@aucklanduni.ac.nz}

\author{Zoran Salcic}
\affiliation{%
   \institution{The University of Auckland}
   \country{New Zealand}}
\email{z.salcic@auckland.ac.nz}

\author{Lichao Sun}
\affiliation{%
   \institution{Lehigh University}
   \country{USA}}
\email{lis221@lehigh.edu}

\author{Gillian Dobbie}
\affiliation{%
   \institution{The University of Auckland}
   \country{New Zealand}}
\email{g.dobbie@auckland.ac.nz}

\author{Philip S. Yu}
\affiliation{%
   \institution{University of Illinois at Chicago}
   \country{USA}}
\email{psyu@uic.edu}

\author{Xuyun Zhang}
\affiliation{%
   \institution{Macquarie University}
   \country{Australia}}
\email{xuyun.zhang@mq.edu.au}

\renewcommand{\shortauthors}{H. Hu et al.}

\begin{abstract}
Machine learning (ML) models have been widely applied to various applications, including image classification, text generation, audio recognition, and graph data analysis. However, recent studies have shown that ML models are vulnerable to membership inference attacks (MIAs), which aim to infer whether a data record was used to train a target model or not. MIAs on ML models can directly lead to a privacy breach. For example, via identifying the fact that a clinical record that has been used to train a model associated with a certain disease, an attacker can infer that the owner of the clinical record has the disease with a high chance. In recent years, MIAs have been shown to be effective on various ML models, e.g., classification models and generative models. Meanwhile, many defense methods have been proposed to mitigate MIAs. Although MIAs on ML models form a newly emerging and rapidly growing research area, there has been no systematic survey on this topic yet. In this paper, we conduct the first comprehensive survey on membership inference attacks and defenses. We provide the taxonomies for both attacks and defenses, based on their characterizations, and discuss their pros and cons. Based on the limitations and gaps identified in this survey, we point out several promising future research directions to inspire the researchers who wish to follow this area. This survey not only serves as a reference for the research community but also provides a clear description for researchers outside this research domain. To further help the researchers, we have created an online resource repository, which we will keep updated with future relevant work. Interested readers can find the repository at \url{https://github.com/HongshengHu/membership-inference-machine-learning-literature}.
\end{abstract}


\maketitle

\section{Introduction}
\label{sec01::intro}
Machine learning (ML) has achieved tremendous results for various learning tasks, including image recognition~\cite{he2016deep}, natural language processing~\cite{devlin2018bert}, graph data applications \cite{kipf2016semi}, as well as advanced applications such as brain circuits analysis \cite{litjens2017survey}, healthcare analysis \cite{miotto2018deep}, and functionality of mutations in DNA \cite{xiong2015human}. Besides the powerful computational resources, the availability of large datasets is another key factor contributing to the success of ML~\cite{bengio2021deep}. As datasets can contain individuals’ private information such as user speech, images, and medical records, it is essential that ML models should not leak privacy sensitive information about their training data. However, recent studies~\cite{zhang2021understanding,song2017machine,carlini2019secret} have shown that ML models are prone to memorizing information of training data, making them vulnerable to several privacy attacks such as model extraction attacks~\cite{tramer2016stealing}, attribute inference attacks (also known as model inversion attacks)~\cite{fredrikson2015model}, property inference attacks~\cite{ganju2018property}, and membership inference attacks~\cite{shokri2017membership}. Model extraction attacks aim to duplicate the functionality of an ML model, i.e., an attacker tries to construct another model whose predictive performance is similar to the target ML model. Unlike model extraction attacks targeting the ML model, the attacker of an attribute inference attack, property inference attack or membership inference attack focuses on inferring private information of the training data. More specifically, attribute inference attacks aim to infer sensitive attributes of a target data record given the output of a model and the information about the non-sensitive attributes. Property inference attacks aim to infer the global property of the training dataset. For example, given a malware classifier model whose training data consists of the execution traces of malicious and benign software, property inference attacks infer the property of the testing environment, which can be viewed as a property of the entire training dataset. Membership inference attacks (MIAs), which are also the focus of this paper, aim to infer members of the training dataset of an ML model. We discuss MIAs in detail subsequently.

Membership inference attacks (MIAs) on ML models aim to infer whether a data record was used to train a target ML model or not. MIAs can raise severe privacy risks to individuals. For example, by identifying the fact that a clinical record has been used to train a model associated with a certain disease,  MIAs can infer that the owner of the clinical record has the disease with a high chance. A recent report~\cite{tabassi2019taxonomy} published by the  National Institute of Standards and Technology (NIST) specifically mentions that an MIA determining an individual was included in the dataset used to train the target model is a confidentiality violation. Moreover, such privacy risks caused by MIAs can lead to commercial companies who wish to release machine learning as a service (MLaaS) to violate privacy regulations. For example, Veale et al.~\cite{veale2018algorithms} mention that MIAs on ML models increase their risks of being classified as private personal information under the General Data Protection Regulation (GDPR)~\cite{wikipedia_gdpr}. The concept of MIAs is firstly proposed by Homer et al. in~\cite{homer2008resolving} where they demonstrate an attacker can leverage the published statistics about a genomics dataset to infer the presence of a particular genome in this dataset. Also, recent papers~\cite{pyrgelis2018knock,pyrgelis2020measuring} have shown the feasibility of MIAs on location data. Besides the MIAs on such databases, Shokri et al.~\cite{shokri2017membership} proposed the first MIAs on several classification models in the context of ML. They demonstrate that an attacker can identify whether a data record was used to train a neural network based classifier or not, solely based on the prediction vector of the data record (which is also known as black-box access to the target ML model). Since then, there have been an increasing number of studies that investigate MIAs on various ML models, including regression models~\cite{gupta2021membership}, classification models~\cite{shokri2017membership}, generation models~\cite{hayes2019logan}, and embedding models~\cite{song2020information}. Meanwhile, a large body of work proposes different membership inference defenses from different perspectives to defend against MIAs while preserving the utility of the target ML models. 

Given the importance of data privacy protection and the successful applications of ML models in various domains, both academia and industry are interested in the privacy of ML models. In this paper, we contribute the first study summarizing different membership inference attacks and defenses on ML models, and establish taxonomies based on various criteria for the relevant research communities. There are many surveys that summarize different attacks on ML models~\cite{liu2021machine,yin2021comprehensive,rosenberg2021adversarial,serban2020adversarial,liu2021trustworthy,papernot2016towards,sun2018adversarial,rigaki2020survey,de2020overview,jere2020taxonomy,mireshghallah2020privacy}. Among them, a line of surveys~\cite{rosenberg2021adversarial,serban2020adversarial,sun2018adversarial} focuses on adversarial attacks~\cite{szegedy2013intriguing} which can lead to severe security risks in critical ML application domains such as self-driving cars, health care, and cybersecurity. For example, in~\cite{rosenberg2021adversarial} the authors comprehensively summarize and shed a light on the risks of the latest studies on adversarial attacks in the domain of cybersecurity. Another line of surveys~\cite{liu2021machine,yin2021comprehensive,rigaki2020survey,de2020overview,mireshghallah2020privacy} focuses on privacy attacks, which can breach personal information that violates the privacy of an individual. These survey papers either investigate the privacy issues in a specific paradigm like federated learning~\cite{jere2020taxonomy}, or provide general discussions~\cite{de2020overview} on different privacy attacks such as attribute inference attacks, property inference attacks, and membership inference attacks. The existing surveys of privacy attacks~\cite{liu2021machine,rigaki2020survey,de2020overview,yin2021comprehensive,jere2020taxonomy,mireshghallah2020privacy} have mentioned MIAs with basic introductions to the concepts and shallow discussions of the methods. In contrast, our survey on MIAs differs from them significantly in scope and depth. Instead of covering all the privacy attacks, we focus only on MIAs given that they have emerged recently and are of great interest to the research community due to their high likelihood of compromising the privacy of training data. Unlike the existing reviews that select a very limited number of publications related to MIAs, e.g., only eight references are included in~\cite{liu2021machine}, we conduct a comprehensive search and include more than 100 related works in this survey. Our survey work offers deeper discussion on the concepts, theories, methods, categorization with taxonomies, and visions of future research directions. The main contributions of this paper are:
\begin{enumerate}
     
    \item \textbf{Comprehensive Review.} To the best of our knowledge, this is the first work to provide a comprehensive review of membership inference attacks and defenses on ML models. We summarize most, if not all, the published and pre-print works (over 100 papers) before September 2021. In this work, we establish novel taxonomies for membership inference attacks and defenses, respectively, according to various criteria.
    
    \item \textbf{Taxonomies of Membership Inference on ML Models.} There are already over a hundred papers published in this domain. A list of all papers can help but is not good enough for readers to quickly understand the similarity and differences among membership inference attacks and among membership inference defenses.
    To this end, we categorize all existing works of MIAs based on different target ML models, adversarial knowledge, attack methods, training algorithms, and task domains, respectively. For membership inference defenses, we categorize all existing works based on different techniques. More details of the taxonomies are given in Fig.~\ref{taxonomy::attack_category} and Fig.~\ref{taxonomy::defense_category}.
    
    \item \textbf{Challenges and Future Directions.} Membership inference attacks on machine learning models is an active and ongoing area of research. Based on the literature reviewed, we have discussed the challenges yet to be solved and proposed several promising future directions for membership inference attacks and membership inference defenses, respectively, to inspire interested readers to explore this field in more depth.
    
    \item \textbf{Datasets and Metrics.} To help researchers conduct empirical studies on membership inference attacks and defenses, we summarize most, if not all, the datasets and metrics that have been used in previous work. This aims to pave the way for the community to build a good benchmark in this area for future empirical analysis and in-depth technical understanding.
    
    \item \textbf{Online Updating Resource.} We create an open-source repository\footnote{\url{https://github.com/HongshengHu/membership-inference-machine-learning-literature}} that includes most, if not all, the relevant work. This repository provides all paper links and released code links to help researchers interested in this area. As a small number of surveyed papers are only available in pre-print, authors are welcome to update us when the full publication information becomes available. We will keep updating the repository with new work in this domain in the future. We hope this open-source repository can shed light on future research about membership inference analysis on ML models.
    
\end{enumerate}

The rest of the paper is organized as follows. Section~\ref{sec02::preli} introduces ML preliminaries. In Section~\ref{sec03::attacks}, we introduce the existing attack approaches and provide taxonomies to categorize the released papers. In Section~\ref{sec04::why}, we discuss why MIAs can work on ML models. Section~\ref{sec05::defense} provides taxonomies for membership inference defenses. Section~\ref{sec06::resources} summarizes datasets, metrics and open-source implementation of popular approaches. We discuss the challenges and propose the future directions in Section~\ref{sec07::future_directions}. We conclude this paper in Section~\ref{sec08::conclusion}.

\section{Preliminaries about Machine Learning}
\label{sec02::preli}
To help the audience understand the context of machine learning where membership inference attacks are performed, we introduce the basic preliminaries of machine learning. It is worth noting that this is not a comprehensive introduction, and interested readers can refer to \cite{mitchell1997machine,goodfellow2016deep,abukmeil2021survey} for a systematic introduction. 

Machine learning (ML) is the study of computer algorithms that improve automatically through experience and by learning from data~\cite{mitchell1997machine}. Generally, ML algorithms can be divided into two categories, i.e., supervised learning and unsupervised learning, depending on the information provided by the training data and the different learning tasks.  

\textbf{Supervised Learning.} A supervised ML model aims to learn a general rule that maps inputs to outputs from a labeled dataset~\cite{russell2002artificial}. Let $D_{train}=\{(\bm{x}^{(n)}, y^{(n)})\}_{n=1}^{N}$ be a training dataset, where $N$ is the number of data instances, $\bm{x}$ is a feature vector, and $y$ is a label. An ML model is a function $f(\bm{x};\bm{\theta})$ that takes as input $\bm{x}$ and outputs $y=f(\bm{x};\bm{\theta})$, where $\bm{\theta}$ are parameters that are learned from $D_{train}$. When $y$ is discrete, the learning task of $f(\bm{x};\bm{\theta})$ is called classification. When $y$ is continuous, the learning task of $f(\bm{x};\bm{\theta})$ is called regression. 

\textbf{Training Supervised ML Models.} A well-trained supervised ML model should have a small expectation loss for the data itworks on. However, as we do not know the true distribution of data, we cannot calculate the model's expected risk. A realistic approach to train supervised ML models is Empirical Risk Minimization (ERM)~\cite{vapnik1992principles}. The core idea is to measure the model's performance on a known training dataset. For a given dataset $D_{train}$, ERM tries to find the parameters 
$\bm{\theta^{*}}$ that minimize the following objective function:
\begin{equation}
\min \mathcal{R}_{{D_{train}}}(\theta)=\frac{1}{N} \sum_{n=1}^{N} \mathcal{L}(y^{(n)}, f(\bm{x^{(n)}} ; \theta)),
\end{equation}
where $\mathcal{L}(\cdot , \cdot)$ is a loss function. An iterative optimization algorithm called \textit{stochastic gradient descent} (SGD)~\cite{saad1998online} is usually used to find the best parameters $\theta^{*}$. The SGD algorithm follows:
\begin{equation}
\theta_{t+1}=\theta_{t}-\alpha \frac{\partial \mathcal{R}_{\mathcal{D}}(\theta)}{\partial \theta},
\end{equation}

\begin{equation}
\frac{\partial \mathcal{R}_{\mathcal{D}}(\theta)}{\partial \theta}= \frac{1}{K} \sum_{n=1}^{K} \frac{\partial \mathcal{L}\left(y^{(n)}, f\left(\boldsymbol{x}^{(n)} ; \theta\right)\right)}{\partial \theta},
\end{equation}
where $K$ is the size of a small batch, $\theta_{t}$ are iterative parameters in the $t_{th}$ time and $\alpha$ is the learning rate. Training is finished when the model converges to a local minimum where the gradient is close to zero. 

\textbf{Unsupervised Learning.} An unsupervised ML model aims to extract features and patterns from unlabeled data or labeled data without access to the labels~\cite{hinton1999unsupervised}. Recently, generative models, which aim to learn how to generate samples from the underlying data distribution, are gaining increasing attention as a typical unsupervised learning method. There are two typical generative models, Generative Adversarial Networks (GANs)~\cite{goodfellow2014generative} and Variational Autoencoders (VAEs)~\cite{kingma2013auto}.  

\textbf{Training Unsupervised ML Models.} We briefly introduce how to train GANs and VAEs because current MIAs on unsupervised learning models mainly target GANs and VAEs.

A GAN consists of two competing neural network modules, a generator $\mathcal{G}$ and a discriminator $\mathcal{D}$, which are trained to compete against each other. The generator takes the latent variable $\bm{z}$ and generates samples $\mathcal{G}_{\theta_{\mathcal{G}}}(\bm{z})$ that approximate the data distribution of $D_{train}$. The discriminator receives samples from $D_{train}$ and the generated samples, and it is trained to learn the difference between them. The discriminator essentially is a binary classifier which determines whether $\bm{x}$ was taken from $D_{train}$ or $\mathcal{G}$. After training, $\mathcal{G}$ can receive different $\bm{z}$ and generate synthetic samples. As both the generator and discriminator are neural networks, SGD is usually used for training GANs. SGD tries to find the parameters $\theta^{*}_{\mathcal{G}}$ of GANs following the objective function:
\begin{equation}
\begin{array}{rl}
\min _{\theta_{\mathcal{G}}} \max _{\theta_{\mathcal{D}}} & \mathbb{E}_{\bm{x} \sim P_{\text {data }}}\left[\log \left(\mathcal{D}_{\theta_{\mathcal{D}}}(\bm{x})\right)\right]+ 
 \mathbb{E}_{\bm{z} \sim P_{\bm{z}}}\left[\log \left(1-\mathcal{D}_{\theta_{\mathcal{D}}}\left(\mathcal{G}_{\theta_{\mathcal{G}}}(\bm{z})\right)\right)\right],
\end{array}
\end{equation}
where $\theta_{\mathcal{G}}$ and $\theta_{\mathcal{D}}$ are parameters of the generator and the discriminator in GANs, $P_{\text{data}}$ is the distribution of $D_{train}$, while $P_{\bm{z}}$ is the distribution of the latent variable $\bm{z}$.

VAEs aim to find an approximate posterior distribution function $q(\bm{z}|\bm{x})$ that parameterizes the latent distribution $P_{\bm{z}}$ according to the input data~\cite{kingma2013auto}. A VAE consists of an encoder and a decoder. The encoder maps data into a latent space, while the decoder maps the encoded latent representation back to the data space.  At the beginning, the prior distribution of the latent variable $P_{z}$ is defined as a unit normal distribution. Accordingly, the encoder and decoder are trained jointly such that the output of the decoder minimizes a reconstruction error between the parametric posterior and the true posterior measured by the Kullback-Leibler (KL) divergence~\cite{kullback1997information}. Formally, to train VAEs, SGD tries to find the parameters $\theta^{*}_{\text{de}}$ following the objective function:
\begin{equation}
\min _{\theta_{\text{en}}, \theta_{\text{de}}}-\mathbb{E}_{q_{\theta_{\text{en}}}(\bm{z} \mid \bm{x})}\left[p_{\theta_{\text{de}}}(\bm{x} \mid \bm{z})\right]+KL\left(q_{\theta_{\text{en}}}(\bm{z} \mid \bm{x}) \| P_{\bm{z}}\right),
\end{equation}
where $q_{\theta_{\text{en}}} (\bm{z} \mid \bm{x})$ and $p_{\theta_{\text{de}}} (\bm{x} \mid \bm{z})$ are the encoder and the decoder, and $\theta_{\text{en}}$ and $\theta_{\text{de}}$ are their parameters, $K L(\cdot \| \cdot)$ is the KL divergence and $P_{\bm{z}}$ is the distribution of $\bm{z}$.

\section{Membership Inference Attacks on Machine Learning Models}
\label{sec03::attacks}
In this section, we first give a general definition of MIAs on ML models, and then introduce adversarial knowledge, attack approaches, and target models of MIAs.

\begin{figure}[!t]
\centering
\includegraphics[width=0.65\linewidth]{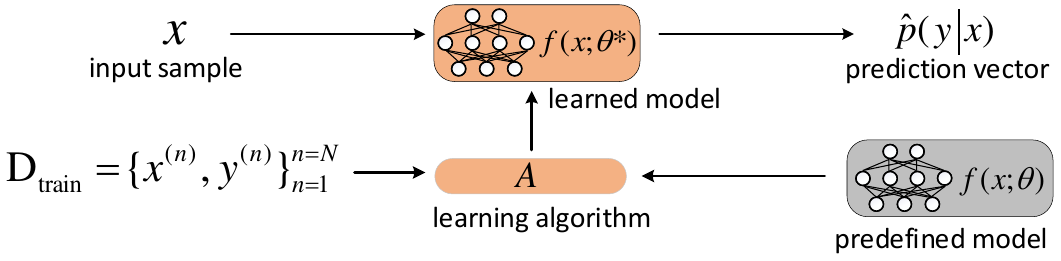}
\caption{A typical deep learning process for classification models.}
\label{fig::ml_process}
\end{figure}

\subsection{Definition of Membership Inference Attacks}
In order to better illustrate the definition of MIAs, we introduce a typical learning process of an ML model. Fig.~\ref{fig::ml_process} shows the typical learning process of a deep neural network (DNN) classifier. We use a learning algorithm $\mathcal{A}$ to train a predefined classifier $f(\bm{x};\theta)$ using dataset $D_{\textrm{train}}=\{(\bm{x}^{(n)}, y^{(n)})\}_{n=1}^{N}$. Once the training process is finished, the learned model $f(\bm{x};\theta^{*})$ can be used to make predictions for unseen data. The definition of MIAs on ML models is as follows: Given an exact input $\bm{x}$ and access to the learned model $f(\bm{x};\theta^{*})$, an attacker infers whether $\bm{x} \in D_{\textrm{train}}$ or not.

\subsection{Adversarial Knowledge}
The attacker can receive different amounts of information to attack ML models. In this section, we first introduce adversarial knowledge, and then introduce black-box and white-box MIAs.

There are two kinds of knowledge that are useful for an attacker to implement MIAs on ML models, i.e., knowledge of training data and knowledge of the target model. Knowledge of training data refers to the distribution of training data. In most settings of MIAs, the distribution of training data is assumed to be available to an attacker of MIAs. This means the attacker can obtain a shadow dataset containing data records from the same data distribution as the training records. This assumption is reasonable because the shadow dataset can be obtained by statistics-based synthesis when the data distribution is known and model-based synthesis when the data distribution is unknown~\cite{shokri2017membership}. To conduct a non-trivial MIA, it is often assumed that the shadow dataset and the training dataset are disjoint. Knowledge of the target model refers to how the target model is trained (i.e., the learning algorithm) and the target model's architecture and learned parameters. Based on adversarial knowledge, we can characterize the dangerous levels of existing attacks.

\textbf{White-box Attack.} Under this setting, an attacker can get all information and use it to attack a target ML model. The information includes the distribution of training data, how the target model is trained, and the architecture and the learned parameters of the target model.

\textbf{Black-box Attack.} In this case, an attacker can only have black-box access to a target model. The attacker is given information limited to training data distribution and black-box queries on the target model. For example, the attacker queries the target classifier (if the target model is a classification model) and only gets prediction output of the input record.

\begin{figure}[!t]
\centering
\includegraphics[width=0.65\linewidth]{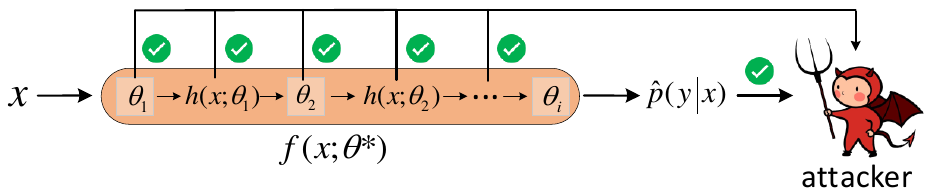}
\caption{Overview of white-box membership inference attacks.}
\label{fig::white-box-attacks}
\end{figure}

\begin{figure}[!t]
\centering
\includegraphics[width=0.65\linewidth]{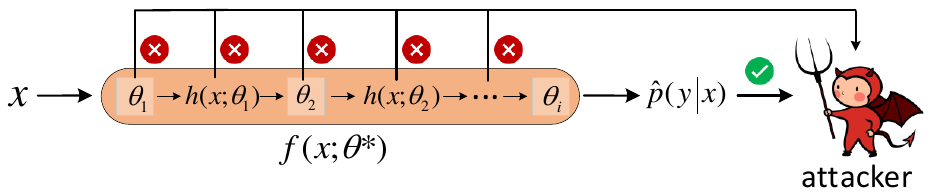}
\caption{Overview of black-box membership inference attacks.}
\label{fig::black-box-attacks}
\end{figure}

We refer to Nasr et al.'s way of depicting white-box and black-box MIAs~\cite{nasr2019comprehensive}, and draw Fig.~\ref{fig::white-box-attacks} and Fig.~\ref{fig::black-box-attacks} to illustrate the concepts of these two types of attacks and their differences for better visual understanding. Fig.~\ref{fig::white-box-attacks} and Fig.~\ref{fig::black-box-attacks} show white-box and black-box MIAs on a target ML model respectively (assuming the target model is a DNN classification model). In both figures, a green tick indicates availability, and the red cross indicates unavailability. In the white-box setting, an attacker has full access to the target classifier and obtains all information, including the learned parameters of the classifier, the prediction vector, and intermediate computations of internal layers when querying an input record. However, in the black-box setting, the attacker has black-box access to the classifier and only receives the prediction vector of the input record. When the target models are classifiers, based on the different information provided by the prediction vector, black-box MIAs can be further divided into three categories as shown in Table~\ref{table::black-box-attacks}.

Compared to white-box MIAs, an attacker of black-box MIAs gets limited information to attack the target ML model. However, if the black-box MIAs can work, they would be more dangerous compared with white-box MIAs because the attacker can breach the membership privacy with limited knowledge. The black-box MIAs on classification models where an attacker is only given the knowledge of a prediction label is the most dangerous attack among all MIAs because the attacker can attack the model with the most limited knowledge. Most existing works study black-box MIAs on classification models with the knowledge of full confidence scores. There are many opportunities to study white-box attacks and black-box attacks with different levels of knowledge.

\begin{table*}
\centering
\caption{Three types of black-box membership inference attacks based on different information provided by the prediction vector.}
\label{table::black-box-attacks}
\resizebox{\linewidth}{!}{%
\begin{tabular}{l|m{13cm}}
\toprule
\textbf{Prediction Output} & \textbf{Description}\\
\hline
Full confidence scores  & The attacker queries an input record and obtains all confidence scores returned by the target classifier. Based on this, the attacker can obtain the predicted label of the target record. Thus, the attacker can further calculate the input's prediction loss (e.g., cross-entropy loss) because this attacker knows the predicted label.\\
\hline
Top-K confidence scores & The attacker queries an input record and obtains only top-K confidence scores returned by the target classifier. For example, the attacker only receives the probabilities of the most likely three classes (assuming the total number of classes is much larger than three). \\
\hline
Prediction label only & The attacker queries an input record and obtains only the predicted label returned by the target classifier. In this case, the attacker is given the most limited knowledge.\\
\bottomrule
\end{tabular}
}
\end{table*}

\subsection{Membership Inference Attack Approaches}
Machine learning (ML) models such as DNNs are often overparameterized, which means that they have sufficient capacity to memorize information about their training dataset~\cite{song2017machine,zhang2021understanding,carlini2019secret,murakonda2020ml}. Moreover, the training datasets are finite in size, and ML models are trained over multiple (often tens to hundreds) epochs on the same instances repeatedly. Consequently, ML models exhibit a different behavior on training data records (i.e., members) versus test data records (i.e., non-members), and also in the model's parameters which store statistically correlated information about specific data records
in their training dataset~\cite{murakonda2020ml,shokri2017membership,nasr2019comprehensive}. For example, a classification model would classify a training data record to its true class with a high confidence score while classifying a test data record to its true class with a relatively small confidence. These different behaviors of ML models enable an attacker of MIAs to build attack models to distinguish members from non-members of the training dataset. Based on the construction of the attack model, there are two major types of MIA approaches, i.e., binary classifier-based attack approaches and metric based attack approaches.

\subsubsection{Binary Classifier Based Membership Inference Attacks}

Essentially, a binary classifier based MIA involves training a binary classifier, which can distinguish a target model's behavior of its training members from the non-members. The challenge is how to train such a binary classifier. An effective technique called \textbf{shadow training} proposed by Shokri et al.~\cite{shokri2017membership} is the first and perhaps the most widely used approach for training a binary classifier based MIA. The main idea is an attacker can create multiple shadow models to mimic the behavior of the target model, because the attacker is assumed to know the structure and the learning algorithm of the target model. For these shadow models, the attacker has their training datasets and test datasets, and thus can construct a dataset containing features and ground truth of membership of the training and test data records. Based on the constructed dataset, the attacker can train the binary classifier-based attack model. 

\begin{figure*}[!t]
\centering
\includegraphics[width=\linewidth]{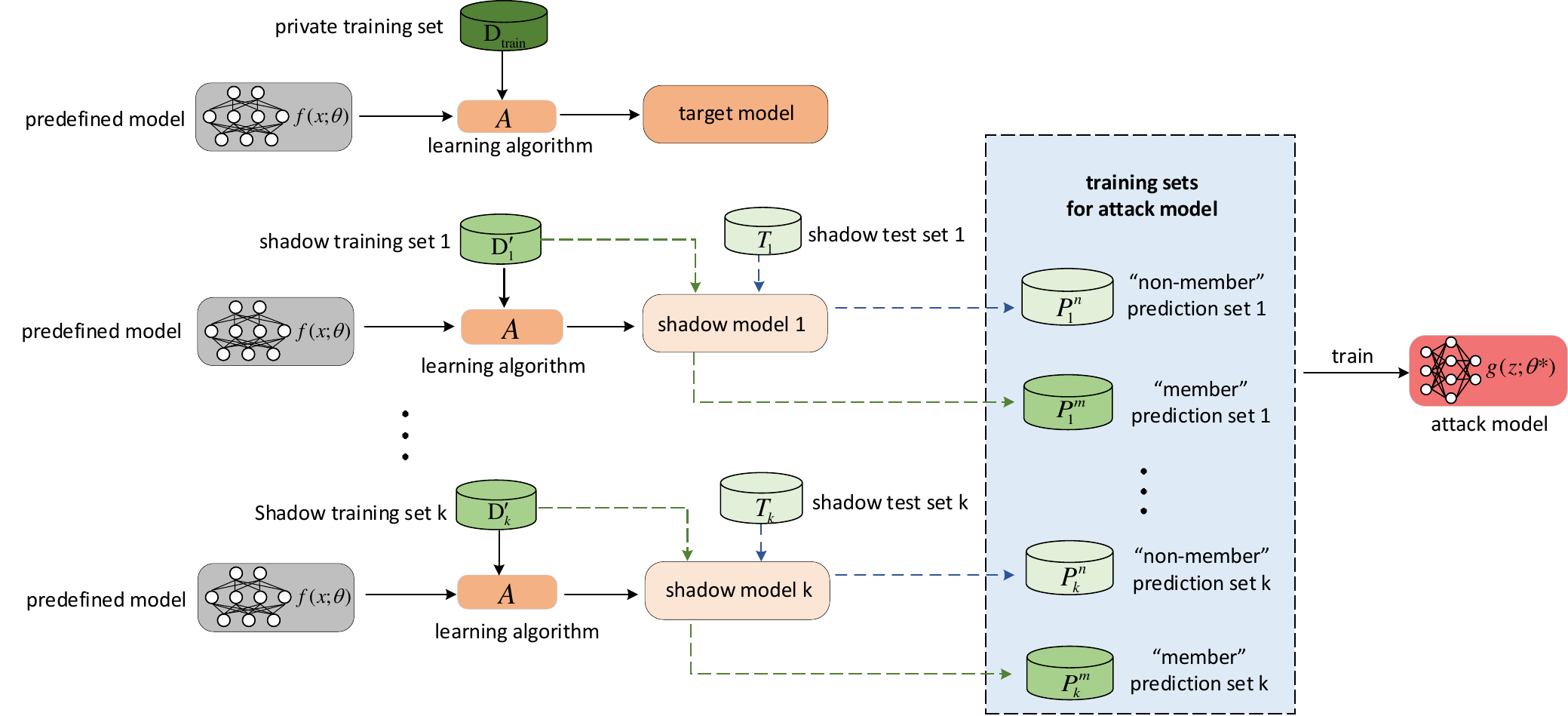}
\caption{Overview of the shadow training technique.}
\label{fig::shadow_training}
\end{figure*}

Fig.~\ref{fig::shadow_training} shows how to use shadow training to train a binary classifier-based attack model to implement MIAs on classification models. $D_{\textrm{train}}$ is a private training dataset, which is used for training the target classifier using the learning algorithm $\mathcal{A}$. $D^{\prime}_1, \cdots, D^{\prime}_k$ are shadow training datasets that are disjoint from the private training dataset $D_{\textrm{train}}$. Each shadow training dataset contains data records coming from the same data distribution as the training members in $D_{\textrm{train}}$, because the attacker is assumed to know the distribution of training data. The attacker first trains $k$ shadow models using shadow training datasets $D^{\prime}_1, \cdots, D^{\prime}_k$ and the learning algorithm $\mathcal{A}$. Each shadow model is trained in such a way to mimic the behavior of the target model. $T_1, \cdots, T_k$ are shadow test datasets which are disjoint from $D^{\prime}_1, \cdots, D^{\prime}_k$. The more shadow models, the more accurate the attack model can be because more shadow models can provide more training fodder for the attack model~\cite{shokri2017membership}. When the shadow models finish training, the attacker queries each of the shadow models using its shadow training dataset and shadow test dataset to obtain the outputs, which are prediction vectors of each data record. For each shadow model, the prediction vector of each record in the shadow training dataset is labeled ``member'' and the prediction vector of each record in the shadow test dataset is labeled ``non-member''. Thus, the attacker can construct $k$ ``member'' datasets and $k$ ``non-member'' datasets, which jointly consist of the training datasets for the attack model. Finally, the problem of recognizing the complex relationship between members and non-members of the training dataset is converted into a binary classification problem. Because binary classification is a standard ML task, the attacker can use any state-of-the-art ML framework to build the attack model.

The shadow training technique can be used for training both white-box and black-box attack models. In both MIAs, the training procedure of the attack models is the same as shown in Fig.~\ref{fig::shadow_training}. However, because adversarial knowledge available for an attacker of black-box and white-box MIAs is different, the attacker can collect different amounts of information about the training members and non-members under the different settings. In the black-box setting, the attacker only has black-box access to the target model, which means the attacker can only receive the prediction vector of an arbitrary input record when querying the target model. Thus, when querying the shadow models using their own shadow training datasets and test datasets, the attacker only collects the prediction vectors of each data record. However, in the white-box setting, the attacker has full access to the target model, which means the attacker can observe the intermediate computations at hidden layers and the prediction vector of an arbitrary input record. Thus, in the white-box setting, when querying the shadow models, the attacker can collect prediction vectors in addition to the intermediate computations of each data record. Compared to black-box MIAs, the attacker of white-box MIAs gets much more information to build the attack model. Next, we show more details of how an attacker constructs the attack model in both settings.

\begin{figure}[!t]
\centering
\subfloat[Binary classifier based black-box MIAs.\label{fig::binary-black-box-attack}]{%
       \includegraphics[width=0.45\textwidth]{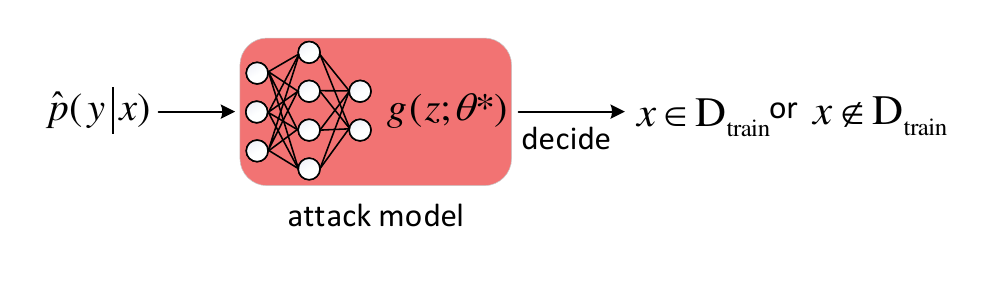}}
    \vspace{10pt}
  \subfloat[Binary classifier based white-box MIAs.\label{fig::binary-white-box-attack}]{%
        \includegraphics[width=0.45\textwidth]{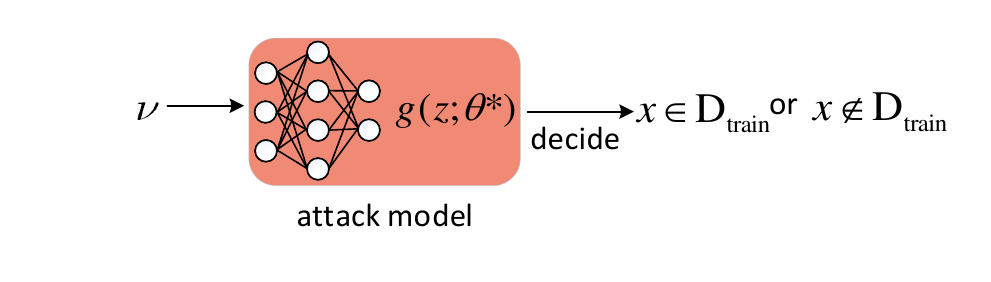}}

\caption{Overview of binary classifier-based attack models in black-box and white-box settings. In the membership inference phase, the black-box attack model only takes the prediction vector $\hat{p}(y \mid \bm{x})$ as input and outputs the membership status of the data record. However, the white-box attack model can take the flat vector $\bm{\nu}$ containing much more information of the data record as input and outputs its membership status.}
\label{fig::binary-mias}
\end{figure}

\noindent \textbf{Binary Classifier Based MIA in Black-box Setting.} Datasets $P^{\textrm{m}}_1,\cdots,P^{\textrm{m}}_k$ are ``member'' datasets which contain prediction vectors of the data records in the shadow training datasets. Datasets $P^{\textrm{n}}_1,\cdots,P^{\textrm{n}}_k$ are ``non-member'' datasets which contain prediction vectors of the data records in the shadow test datasets. We denote a prediction vector as $\hat{p}\left(y \mid \bm{x}\right)$, ``member'' as $1$, and ``non-member'' as $0$. Then, each ``member'' dataset and ``non-member'' dataset is represented as follows:

\begin{equation}
P_{i}^{\textrm{m}}=\left\{\hat{p}\left(y \mid \bm{x}^{(t)}\right), 1\right\}_{t=1}^{N_{i}^{\textrm{m}}},
\end{equation}

\begin{equation}
P_{i}^{\textrm{n}}=\left\{\hat{p}\left(y \mid \bm{x}^{(t)}\right), 0\right\}_{t=1}^{N_{i}^{\textrm{n}}}.
\end{equation}
For an binary classifier $g(\bm{z};\theta)$ (assuming the classifier is a DNN classifier), the attacker uses an SGD algorithm to find parameters $\theta^{*}$ that minimize the following objective function:

\begin{equation}
\mathcal{R}(\theta)=\frac{1}{N} \sum_{n=1}^{N} \mathcal{L} \left({\mathcal{I}}(\bm{x}), g(\hat{p}(y \mid \bm{x}) ; \theta) \right),
\end{equation}
where $N$ is the total number of shadow data records, $\mathcal{L}(\cdot , \cdot)$ is a binary cross entropy loss function, and $I(\cdot)$ is an indicator function as follows:

\begin{equation}
    \mathcal{L}(y,p) = -(y \textrm{log} (p)+(1-y) \textrm{log} (1-p)),
\end{equation}

\begin{equation}
\mathcal{I}(\bm{x}) = \begin{cases}
1 \quad {\textrm{if} \; \bm{x} \in {P^\textrm{m}}}, \\
0 \quad {\textrm{if} \; \bm{x} \notin {P^\textrm{m}}}.
\end{cases}
\end{equation}
After the binary classifier is trained, the attacker can use it as the attack model to implement MIAs on arbitrary data records. Fig.~\ref{fig::binary-black-box-attack} demonstrates a black-box MIA using the trained attack model. The binary classifier $g(\bm{z};\theta^{*})$ takes the prediction vector $\hat{p}(y \mid \bm{x})$ of a data record as input and outputs whether this record is in $D_{\textrm{train}}$ or not.

\noindent \textbf{Binary Classifier Based MIA in White-box Setting.} The attacker can get all the information to implement MIAs in a white-box setting. When querying a shadow model on an input record of the shadow datasets, the attacker can collect the prediction vector $\hat{p}(y \mid \bm{x})$, the intermediate computation $h(\bm{x};\theta_i)$ at each hidden layer, the loss $\mathcal{L}\left(y, \hat{p}(y \mid \bm{x})\right)$, and the gradient of the loss with respect to the parameters of each layer $\frac{\partial \mathcal{L}}{\partial {\theta}_{i}}$ of the input record. In this case, $P^{m}_1,\cdots,P^{m}_k$ are ``member'' datasets which contain the above computations of each data record in the shadow training sets, and $P^{n}_1,\cdots,P^{n}_k$ are ``non-member'' datasets which contain the computations of each data record in the shadow test datasets. The attacker then concatenates all the computations of each data record into a flat vector as follows: 

\begin{equation}
\begin{array}{c}
\bm{\nu} = (\frac{\partial \mathcal{L}}{\partial \theta_{1}}, h\left(\bm{x} ; \theta_{1}\right), \cdots, \frac{\partial \mathcal{L}}{\partial \theta_{i}}, h\left(\bm{x} ; \theta_{i}\right),
\hat{p}(y \mid \bm{x}), \mathcal{L}(y ; \hat{p}(y \mid \bm{x}))).
\end{array}
\end{equation}
Then, each ``member'' dataset and ``non-member'' dataset is represented as follows:
\begin{equation}
P_{i}^{\textrm{m}}=\left\{\bm{\nu}, 1\right\}_{t=1}^{N_{i}^{\textrm{m}}},
\end{equation}

\begin{equation}
P_{i}^{\textrm{n}}=\left\{\bm{\nu}, 0\right\}_{t=1}^{N_{i}^{\textrm{n}}}.
\end{equation}
The structure of the binary classifier-based attack model in the white-box setting is usually different from that in the black-box setting, because the input of the attack model in the two settings is very different. Nevertheless, the attack model is a binary classifier. For a binary classifier $g(\bm{z};\theta)$, the attacker uses SGD to find parameters $\theta^{*}$ that minimize the following objective function:

\begin{equation}
\mathcal{R}(\theta)=\frac{1}{N} \sum_{n=1}^{N} \mathcal{L}({I}(\bm{x} \in P^{\textrm{m}}), g(\bm{\nu} ; \theta)).
\end{equation}
Fig.~\ref{fig::binary-white-box-attack} demonstrates a white-box MIA using the trained attack model. The binary classifier $g(\bm{z};\theta^{*})$ takes the flat vector $\bm{\nu}$ of a data record as input and outputs whether this record is in $D_{\textrm{train}}$ or not.

\subsubsection{Metric Based Membership Inference Attacks}
\label{sec::metric-based-attacks}
Unlike binary classifier based MIAs relying on training a binary classifier to recognize the complex relationship between members and non-members, metric based MIAs are more simple and less computational. Metric based MIAs make membership inference decisions for data records by first calculating metrics on their prediction vectors. The calculated metrics are then compared with a preset threshold to decide the membership status of the data record. Based on different metric options, there are four major types of metric based MIAs, i.e., prediction correctness based, prediction loss based, prediction confidence based, and prediction entropy based attacks. We denote a metric based MIA as $\mathcal{M}(\cdot)$, which codes members as $1$, and non-members as $0$. We introduce the detailed metric based attack approaches as follows. Each approach name follows the reference of the first paper that proposes or uses this attack approach. 

\noindent \textbf{Prediction Correctness Based MIA~\cite{yeom2018privacy}.} An attacker infers an input record $\bm{x}$ as a member if it is correctly predicted by the target model, otherwise the attacker infers it as a non-member. The intuition is that the target model is trained to predict correctly on its training data, which may not generalize well on the test data. The attack $\mathcal{M}_{\textrm{corr}}(\cdot ,\cdot)$ is defined as follows:

\begin{equation}
    {\mathcal{M}_{\textrm{corr}}}(\hat p(y\left| \bm{x} \right.),y) = \mathds{1}(\textrm{argmax} \, \, \hat p(y\left| \bm{x} \right.) = y),
\end{equation}
where $\mathds{1}(\cdot)$ is an indicator function as follows:

\begin{equation}
\mathds{1}(A) = \begin{cases}
1 \quad {\textrm{if} \; \textrm{event} \; A \; \textrm{occurs}}, \\
0 \quad {\textrm{otherwise}}.
\end{cases}
\end{equation}

\noindent \textbf{Prediction Loss Based MIA~\cite{yeom2018privacy}.} An attacker infers an input record as a member if its prediction loss is smaller than the average loss of all training members, otherwise the attacker infers it as a non-member. The intuition is that the target model is trained on its training members by minimizing their prediction loss. Thus, the prediction loss of a training record should be smaller than the prediction loss of a test record. The attack $\mathcal{M}_{\textrm{loss}}(\cdot,\cdot)$ is defined as follows:

\begin{equation}
    {\mathcal{M}_{\textrm{loss}}}(\hat p(y\left| \bm{x} \right.),y) = \mathds{1}(\mathcal{L}( \hat p{(y\left| \bm{x} \right.)};y) \le \tau ).
\end{equation}
where $\mathcal{L}(\cdot)$ is the cross-entropy loss function and $\tau$ is a preset threshold.

\noindent \textbf{Prediction Confidence Based MIA~\cite{salem2019ml}.} An attacker infers an input record as a member if its maximum prediction confidence is larger than a preset threshold, otherwise the attacker infers it as a non-member. The intuition is that the target model is trained by minimizing prediction loss over its training data, which means the maximum confidence score of a training member's prediction vector should be close to $1$. The attack $\mathcal{M}_{\text{conf}}(\cdot)$ is defined as follows:

\begin{equation}
    {\mathcal{M}_{\textrm{conf}}}(\hat p(y\left| \bm{x} \right.)) = \mathds{1}(\textrm{max} \; \hat p{(y\left| \bm{x} \right.)} \ge \tau ).
\end{equation}

\noindent \textbf{Prediction Entropy Based MIA~\cite{salem2019ml}.} An attacker infers an input record as a member if its prediction entropy is smaller than a preset threshold, otherwise the attacker infers it as a non-member. The intuition is that the prediction entropy distributions between training and test data are very different. The target model usually has a larger prediction entropy on its test data than its training data. The entropy of a prediction vector $\hat{p}(y \mid \bm{x})$ is defined as follows:

\begin{equation}
    H(\hat p(y\left| \bm{x} \right.)) =  - \sum\nolimits_i {{p_i}} \log ({p_i}),
\end{equation}
where $p_i$ is the confidence score in $\hat{p}(y \mid \bm{x})$. The attack $\mathcal{M}_{\text{entr}}(\cdot)$ is then defined as follows:

\begin{equation}
    {\mathcal{M}_{\textrm{entr}}}(\hat p(y\left| \bm{x} \right.)) = \mathds{1}(H(\hat p(y\left| \bm{x} \right.)) \le \tau ).
\end{equation}

\noindent \textbf{Modified Prediction Entropy Based MIA~\cite{song2021systematic}.} The authors in \cite{song2021systematic} argue that the existing prediction entropy based MIA does not consider any information about the ground truth label, which might misclassify members and non-members. For example, a totally wrong classification with a probability score of one leads to zero prediction entropy value for an input record. The existing prediction entropy-based MIA will classify the record as a member. However, the record with a totally wrong classification is highly likely a non-member. Thus, they propose a modified prediction entropy metric that can leverage the information of the ground truth label as follows:

\begin{equation}
    {MH}(\hat p(y\left| \bm{x} \right.),y) =  - (1 - {p_y})\log ({p_y}) - \sum\nolimits_{i \ne y} {{p_i}} \log (1 - {p_i}),
\end{equation}
where $p_y$ is the confidence score of the ground truth label. Then, the attack $\mathcal{M}_{\text{Mentr}}(\cdot,\cdot)$ is defined as follows:

\begin{equation}
    {\mathcal{M}_{\textrm{Mentr}}}(\hat p(y\left| \bm{x} \right.),y) = \mathds{1}(MH(\hat p(y\left| \bm{x} \right.);y) \le \tau ).
\end{equation}

\subsection{Membership Inference Attacks on Different ML models}
Since the first work~\cite{shokri2017membership} proposed MIAs on classification models, there has been an increasing number of studies investigating MIAs on classification models as well as other ML models (e.g., generative models). In this section, we select a few pieces of literature to introduce MIAs on specific ML models, including classification models, generative models, embedding models, and regression models. Each of the selected papers either proposes new MIAs or is the first to investigate the membership privacy risks on a specific ML model or under a unique adversarial knowledge setting. 

\noindent \textbf{MIAs on Classification Models. \;} Currently, many of the MIAs focus on classification models. In this case, an attacker aims to infer whether a data instance was used to train a target classifier. Shokri et al.~\cite{shokri2017membership} conducted the pioneering work to propose the first MIA on classification models. They invented a shadow training technique to train a binary classifier-based attack model in a black-box setting. Salem et al.~\cite{salem2019ml} relax two main assumptions of the shadow training technique in \cite{shokri2017membership}, i.e., multiple shadow models and knowledge of the training data distribution. They argue that the two assumptions are relatively strong, which heavily limit the applicable scenarios of MIAs against ML models. They show that even with one single shadow model, the attacker can achieve comparable attack performance compared to using multiple shadow models. They also propose a data transferring attack where a dataset used to train the shadow model is not required to have the same distribution as the target model's private training dataset. Also, the shadow model is not required to have the same structure as the target model. Besides extending existing binary classifier based MIAs in~\cite{shokri2017membership}, they propose two metric based attacks leveraging the highest confidence score and prediction entropy. Yeom et al.~\cite{yeom2018privacy} also propose two metric based MIAs, i.e., the prediction correctness based MIA and the prediction loss based MIA. Compared to binary classifier based MIAs, metric based MIAs are much simpler and have a smaller computation cost. Long et al.~\cite{long2018understanding,long2020pragmatic} investigate MIAs on ML models which are not overfitted to their training data. They propose a generalized MIA that can identify the membership of particular vulnerable records. The intuition is that some records have unique influences on the target model, even when the model is well-generalized. An attacker can exploits the unique influences of particular data records as an indicator of their presence in the training dataset. They show that the vulnerable records can be inferred correctly on well-generalized models even if the gap in training and testing accuracy is smaller than $1\%$.

An attacker of the above MIAs is given full confidence scores of a target record to infer the membership status of the record. Li and Zhang~\cite{li2020label}, and Choquette et al.~\cite{choquette2021label} study MIAs in a more restricted scenario where the target model only provides the predicted label to the attacker. Li and Zhang~\cite{li2020label} propose two label-only MIAs, i.e., a transfer based MIA and a perturbation based MIA. The transfer based MIA aims to construct a shadow model to mimic the target model. The intuition is that if the shadow model is similar enough to the target model, then the shadow model's confidence scores on an input record will indicate its membership. The perturbation based attack aims to add crafted noise to the target record to turn it into an adversarial example. The intuition is that it is harder to perturb a member instance to a different class than a non-member instance. Thus, the magnitude of the perturbation can be used to distinguish members from non-members. Choquette et al.~\cite{choquette2021label} also propose two label-only MIAs, i.e., data augmentation based MIA and decision boundary distance based MIA. For a target record, a data augmentation based attack creates additional data records via different data augmentation strategies. The additional data records are then used to query the target model and the attacker can collect all the predicted labels. The attack intuition is that many models use data augmentation during the training process. Thus, a member record's augmented versions are less likely to change their predicted label. The decision boundary based attack estimates a record's distance to the model's boundary and decides it is a member if its distance is larger than a threshold. The intuition is similar to Li and Zhang's~\cite{li2020label} perturbation based attack. The success of label-only MIAs demonstrate that ML models can be more vulnerable to MIAs than we expect.

While the above MIAs focus on a black-box setting, Nasr et al.~\cite{nasr2019comprehensive} first propose white-box MIAs, where an attacker knows internal parameters of the target model. Their white-box MIAs can be considered as an extension of the binary classifier based MIAs in black-box settings. Compared to black-box MIAs, white-box MIAs try to improve the attack performance by leveraging an input record's intermediate computations through the target model. They use the gradient of an input's prediction loss with regard to the target model's parameters as additional features to infer the membership of the record. The intuition is that the gradients of a training member's loss over the model’s parameters is distinguishable from non-members through the training of the SGD algorithm. However, Leino and Fredrikson~\cite{leino2020stolen} point out that the white-box setting in Nasr et al.'s paper~\cite{nasr2019comprehensive} is too strong, which deviates from most settings of MIAs. Nasr et al.~\cite{nasr2019comprehensive} assume that an attacker knows a significant portion of the target model's private training dataset, while the attacker is often assumed to only have a shadow dataset that is disjoint from the private training dataset. Thus, Leino and Fredrikson propose an effective white-box MIA that does not require any of the target model's training members. The attack intuition is that the membership information can be leaked through a target model's idiosyncratic use of features. Features distributed differently in the training data than in the true distribution can provide evidence for membership. They first build a Bayes-optimal attack assuming the target model is a simple linear softmax model. When the target is a DNN model, they approximate each layer as a local linear model, which is then applied to the Bayes-optimal attack. The attacks on different layers are then combined to compute the final membership decision. 

\noindent \textbf{MIAs on Generative Models. \;} Besides classification models, MIAs are also investigated on generative models. Currently, MIAs on generative models focus on GANs, which is the most popular generative models. We first describe the architecture of a GAN and then introduce specific MIAs on GANs.
\begin{figure}[!t]
\centering
\includegraphics[width=0.60\textwidth]{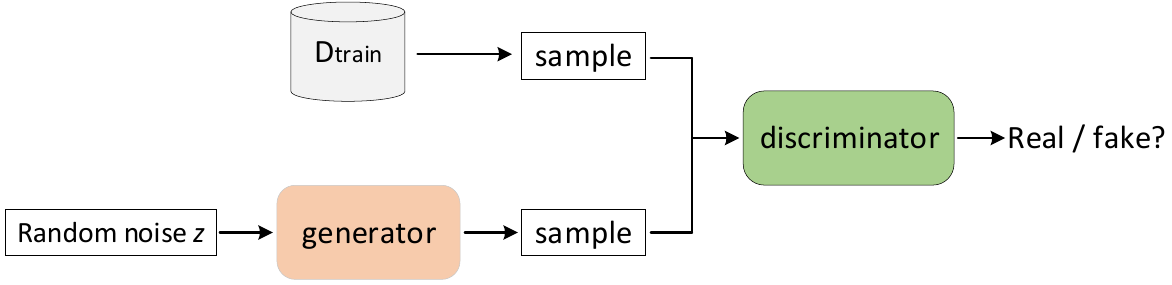}
\caption{Architecture of a generative adversarial network.}
\label{fig::GAN}
\end{figure}

Fig.~\ref{fig::GAN} depicts the architecture of a GAN. A GAN consists of two competing neural network modules, a generator $\mathcal{G}$ and a discriminator $\mathcal{D}$, which are trained to compete against each other. MIAs on generative models aim to identify whether a data record was used to train a generator or not, which is more challenging than MIAs on classification models. Unlike classification models, an attacker of an MIA on a generative model does not obtain confidence scores or prediction labels that are related to the target data record from the generative models. This means the attacker has few clues for implementing MIAs. Moreover, current GAN models often encounter model dropping and mode collapse, leading to a problem of underrepresenting certain data records, which poses additional attack difficulty to the attacker.

Hayes et al.~\cite{hayes2019logan} introduce the first MIA on generative models in both black-box and white-box settings. The attack intuition is that the discriminator of GANs is more confident to output a higher confidence value on training members, as it is trained to learn statistical differences between training data and generated data. In the white-box setting, an attacker puts all data records into the discriminator that will output confidence scores for each record corresponding to the probability of being a member. The attacker sorts these probability values in descending order and picks the first half records as members. In the black-box setting, the attacker collects generated records from the generator and uses them to train a local GAN to mimic the target GAN. After the local GAN has been trained, the attacker implements MIAs using the discriminator of the local GAN following the same attack approach as in the white-box setting. 

Hilprecht et al.~\cite{hilprecht2019monte} propose two MIAs on generative models. One is Monte Carlo integration attack designed for GANs in the black-box setting, and the other is reconstruction attack designed for VAEs in the white-box setting. The Monte Carlo integration attack exploits generated records that are within a small distance of a target record to approximate the probability that this record is a member via Monte Carlo integration~\cite{robert2013monte}. The attack intuition is that the generator of GANs should be able to produce synthetic records that are close to the training members if GANs overfit. The reconstruction attack directly makes use of the loss function of VAEs to calculate the reconstruction error of the target member, and the attack intuition is that training members should have smaller reconstruction errors compared to that of non-members. In addition to MIAs for a single record, Hilpreche et al.~\cite{hilprecht2019monte} introduce the concept of set membership inference where the attacker tries to identify whether a set of records belongs to the training dataset or not. Liu et al.~\cite{liu2019performing} propose co-membership inference, which essentially is the same as the set membership inference proposed by Hilpreche et al.~\cite{hilprecht2019monte}. Liu et al.~\cite{liu2019performing} proposed attack begins with attacking a single target record and then extends to a set of records. For a given record and a generator of the target GAN, the attacker first optimizes a neural network to reproduce the latent variable such that the generator can generate synthetic records nearly matching the target record. The attack intuition is that if a record belongs to the training dataset, the attacker is able to reproduce similar synthetic records close to it. The attacker then measures the L2 distance between the synthetic record and the target record and infers the target record is a member if the distance is smaller than a threshold. This attack method is different from the attacks in~\cite{hilprecht2019monte} because it requires retraining new neural networks for different input data, while the Monte Carlo integration attack and the reconstruction attack in~\cite{hilprecht2019monte} only need fixed synthetic records of the generator. 

Chen et al.~\cite{chen2020gan} propose a generic MIA on generative models which is applicable to all adversarial knowledge settings, from full black-box to full white-box settings. For a target record, the attacker tries to reconstruct a synthetic record that is closest to the target record. The attacker simply finds the synthetic record generated from the generator if possible. Otherwise, the attacker makes use of optimization algorithms to reconstruct the synthetic record. The distance between the reconstructed record and the target record is then used for calculating the probability that this target record is a member. The attack intuition is that the generator should be able to generate more similar samples for members than non-members. To make a more accurate probability estimation, they train a reference GAN with a relevant but disjoint dataset to calibrate the reconstruction error (i.e., the distance). The attacker decides the target record is a member when the calibrated reconstruction error is smaller than a threshold. The MIAs introduced above have been evaluated on state-of-the-art generative models, such as DCGAN~\cite{radford2015unsupervised}, VAEGAN~\cite{larsen2016autoencoding}, PGGAN~\cite{karras2017progressive}, WGANGP~\cite{gulrajani2017improved}, and MEDGAN~\cite{choi2017generating}. Notably, \cite{hilprecht2019monte}, \cite{liu2019performing}, and \cite{chen2020gan} report that VAEs are more susceptible to MIAs compared to GANs because VAEs are more prone to overfiting to their training data than GANs.

\noindent \textbf{MIAs on Embedding Models. \;} Embeddings are mathematical functions that map raw objects (such as words, sentences, and graphs) to real valued vectors with the aim of capturing and preserving important semantic information about the underlying objects. Embeddings have been successfully applied to various domains including natural language processing~\cite{kannan2016smart}, social networks~\cite{grover2016node2vec}, movie feedback~\cite{he2017neural}, and location~\cite{dadoun2019location}. Song and Raghunathan~\cite{song2020information} introduced the first MIAs on word embedding and sentence embedding models. Unlike classification models whose training data consist of input feature vectors and class labels, text embedding models are trained on sequences of words or sentences. Thus, the goal of MIAs on text embedding models is to infer the membership of a sliding window of words or a pair of sentences. Song and Raghunathan~\cite{song2020information} propose a metric based MIA that leverages similarity scores of a sliding window of words or a pair of sentences to infer their membership status. The attack intuition is that words and sentences in the context used for training will be more similar to each other than those of non-members. Mahloujifar et al.~\cite{mahloujifar2021membership} demonstrate that MIAs on embedding models can work even when the embedding layer of the embedding models is not exposed to the attacker. Duddu et al.~\cite{duddu2020quantifying} introduced the first MIA on graph embedding models. They propose a shadow model attack in a black-box setting where the embedding layer of the embedding model is used in graph neural networks (GNN) for node classification problems. The shadow model attack uses the shadow training technique and is essentially a binary classifier. They also propose a confidence score attack in a white-box setting where the attacker can directly access the graph embedding model. The attack intuition is that graph nodes with higher output confidence
prediction are more likely to be members of the graph.

\noindent \textbf{MIAs on Regression Models. \;} Gupta et al.\cite{gupta2021membership} introduced the first MIAs on deep regression models. They focus on age prediction problems where regression models predict a person’s age from their brain MRI scan. Because their work focuses on demonstrating the vulnerability of deep regression models to MIAs, they assume an attack is given white-box access to the target model and has access to some training members of the private training dataset.  The attack model is a binary classifier that leverages features of gradients of parameters, activation, predictions, and labels of target records to infer their membership status.

\subsection{Membership Inference Attacks against Federated Learning}
\label{sec::federated-learning}
\begin{figure}[!t]
\centering
\includegraphics[width=0.5\textwidth]{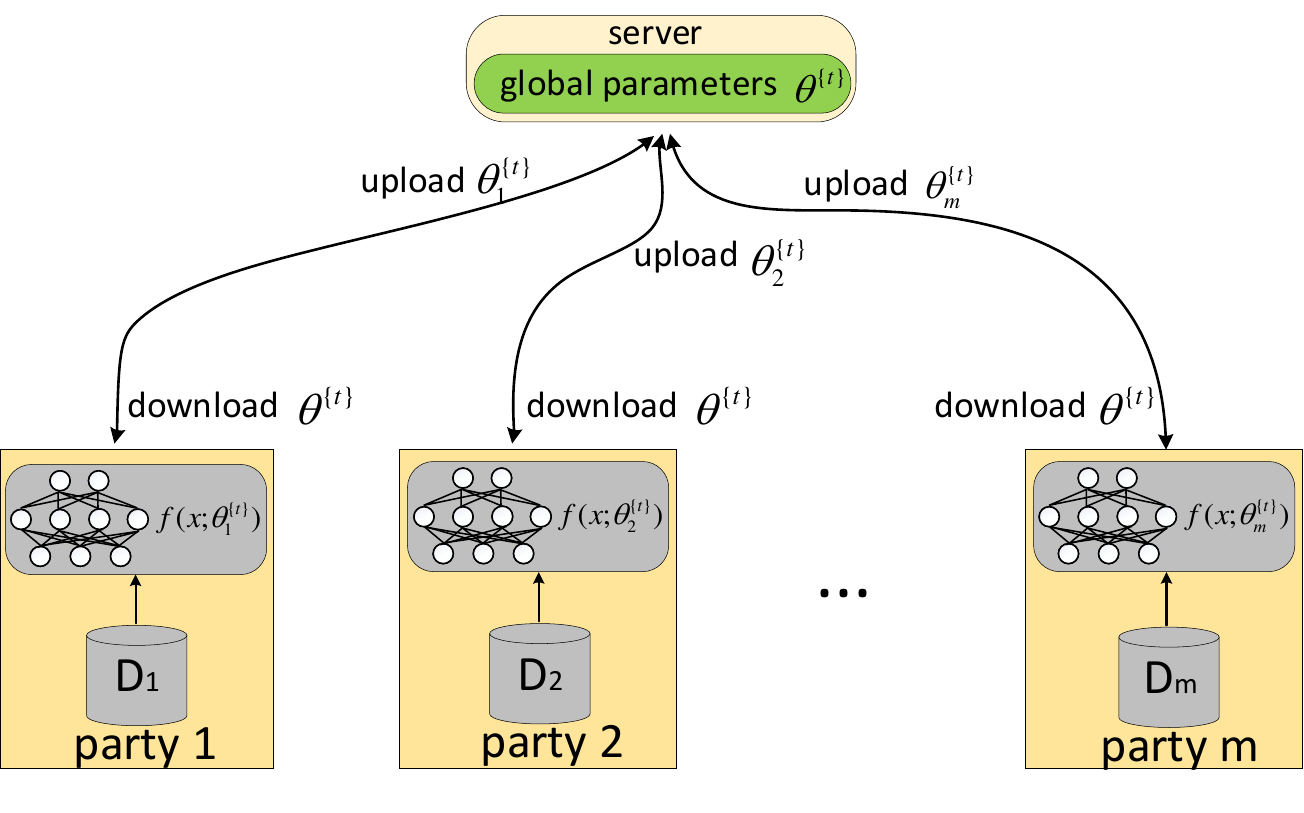}
\caption{Overview of the federated averaging algorithm.}
\label{fig::fed_avg}
\end{figure}
Federated learning (FL) has recently emerged as an alternative to conventional centralized learning where all training data is pooled and a ML model is trained on this joint pool. FL allows multiple parties to jointly train an ML model in an interactive manner. It is an attractive framework for training ML models without direct access to diverse training data owned by different parties, especially for privacy-sensitive tasks~\cite{mcmahan2017communication,mcmahan2018learning}. The above MIAs assume the target ML models are trained in a centralized manner but FL provides interesting new avenues for MIAs. The success of MIAs against FL can shed light on how FL reveals sensitive information and provides an insight that FL may not always provide sufficient privacy guarantees. To better understand membership privacy risks in FL, we first introduce the federated averaging (FedAvg)~\cite{mcmahan2017communication} algorithm, which is the first and perhaps the most widely used FL algorithm. Existing MIAs against FL mainly focus on FedAvg.

Fig.~\ref{fig::fed_avg} shows the FedAvg algorithm~\cite{mcmahan2017communication}. During multiple rounds of communication between server and parties, a central model is trained. At each communication round, the server distributes the current central model to local parties. The local parties then perform local optimization using their own data. To minimize communication, parties might update the local model for several epochs during a single communication round. Next, the optimized local models are sent back to the server, who averages them to allocate a new central model. The performance of the new central model decides the training is either stopped or a new communication round starts. In FL, parties never share data, only their model weights or gradients. 

In the context of FL, the attacker can be either the central server or a certain number of parties who aim to infer whether a data record was used to train the global model. Melis et al.~\cite{melis2019exploiting} introduced the first MIA against FL. They focus on a text classification problem and the target models are recurrent neural networks with a word-embedding layer to transform inputs into a lower-dimensional vector representation via an embedding matrix. The embedding matrix is treated as a parameter of the global model and optimized collaboratively. During training, the embedding layer's gradient is sparse with respect to the input words. This means for a given batch of text, the embedding is updated only with the words that appear in the batch, and the gradients of the other words are zeros. The attacker can observe the non-zero gradients to infer which words occur in the training dataset. Truex et al.~\cite{truex2019demystifying} introduce an MIA against heterogeneous FL where each party trains a local model and only shares confidence scores when making predictions for a new record. They assume different parties have very different datasets, which leads to sufficiently different decision boundaries for different parties. The decision boundary differences reveal the underlying training data, enabling an insider attacker to infer whether a data record is in the local datasets of the other parties.

The attacker of the above MIAs against FL passively follows the FL protocol to infer membership of a data record. However, Nasr et al.~\cite{nasr2019comprehensive} argue that the attacker can also actively tamper with the FL training to achieve better attack performance. They propose an MIA called gradient ascent attack, which intentionally updates the local model parameters in the direction of increasing the loss on a target data record. The attack exploits the fact that SGD optimization updates model parameters in the opposite direction of the gradient of the loss. If the target record is a member, applying the gradient ascent attack will trigger the target model to minimize the loss of this record by descending the model’s gradient and nullify the effect of the attacker’s ascent. However, for a non-member record, the target model will not change their gradient on it explicitly, as this record does not influence the training loss function.

Note that the attacker of an MIA in FL infers whether a data record was used to train the global model, but does not infer whether the data record was used to train a particular local model. This is because MIAs on ML models aim to identify target models' members from non-members of the training dataset. In FL, each party contributes its local training data to jointly train an ML model, and thus the training dataset for the FL model consists of all the local data records. Recently, Hu et al.~\cite{hu2021source} propose source inference attacks that can determine which party owns a training record in FL. They argue that existing MIAs in FL ignore the source of a training member, i.e., the information of the party owning the training member. However, it is essential to explore source privacy in FL beyond membership privacy, because the leakage of such information can lead to further privacy issues. For instance, in the scenario where multiple hospitals jointly train an FL model for the COVID-19 diagnosis, MIAs can only reveal who have been tested for COVID-19, but the further identification of the source hospital where the people are from will make them more prone to discrimination, especially when the hospital is in a high risk region or country~\cite{devakumar2020racism}. They demonstrate that a malicious server in FedAvg~\cite{mcmahan2017communication} can implement source inference attacks effectively and non-intrusively. The intuition of their proposed source inference attacks is that the local model behaves differently on its local training data and the training data of other parties, which enables the malicious server to leverage the prediction loss of local models to steal non-trivial source information of the training members.

\tikzstyle{leaf}=[mybox,minimum height=1em,
fill=hidden-orange!40, text width=20em,  text=black,align=left,font=\tiny,
inner xsep=2pt,
inner ysep=1pt,
]
  
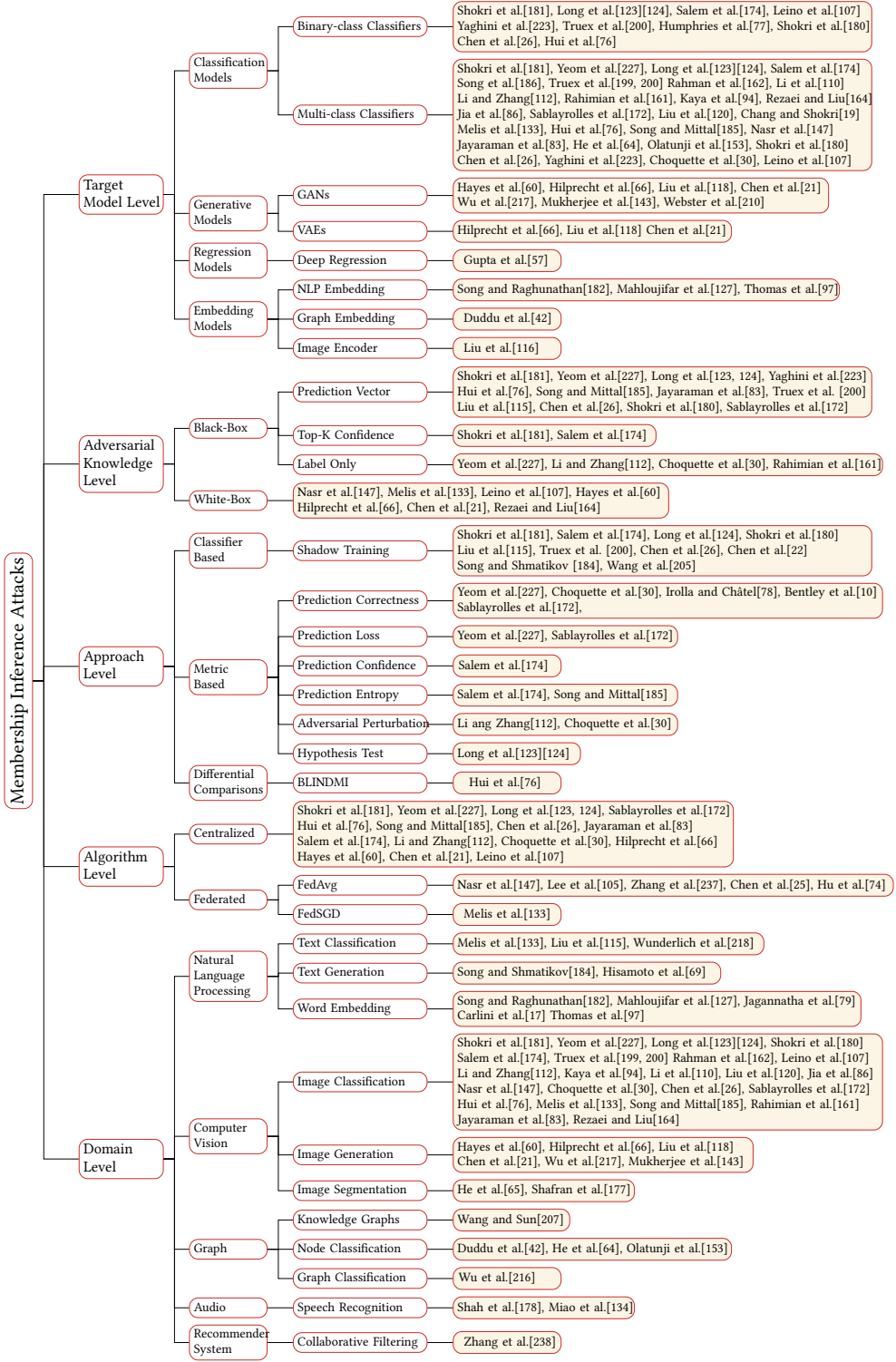
\begin{figure*}[tp]
  \centering
\resizebox{0.95\linewidth}{!}{
\begin{forest}
  forked edges,
  for tree={
  grow=east,
  reversed=true,
  anchor=base west,
  parent anchor=east,
  child anchor=west,
  base=left,
  font=\small,
  rectangle,
  draw=hiddendraw,
  rounded corners,align=left,
  minimum width=2.5em,
s sep=3pt,
inner xsep=2pt,
inner ysep=1pt,
ver/.style={rotate=90, child anchor=north, parent anchor=south, anchor=center},
  },
  where level=1{text width=3.5em,font=\scriptsize,}{},
  where level=2{text width=3.2em,font=\tiny}{},
  where level=3{text width=5.8em,font=\tiny}{},
  [Membership Inference Attacks, ver
    [Target \\ Model Level
        [Classification \\ Models
           [Binary-class Classifiers 
                [Shokri et al.\cite{shokri2017membership}{,} Long et al.\cite{long2018understanding}\cite{long2020pragmatic}{,} Salem et al.\cite{salem2019ml}{,} Leino et al.\cite{leino2020stolen}\\ Yaghini et al.\cite{yaghini2019disparate}{,} Truex et al.\cite{truex2019demystifying}{,} Humphries et al.\cite{humphries2020differentially}{,} Shokri et al.\cite{shokri2021privacy}\\ Chen et al.\cite{chen2020machine}{,} Hui et al.\cite{hui2021practical},leaf,text width=19em
                ]
           ]
           [Multi-class Classifiers 
            [Shokri et al.\cite{shokri2017membership}{,} Yeom et al.\cite{yeom2018privacy}{,} Long et al.\cite{long2018understanding}\cite{long2020pragmatic}{,} Salem et al.\cite{salem2019ml}\\ Song et al.\cite{song2019membership}{,} Truex et al.\cite{truex2019demystifying,truex2019effects} Rahman et al.\cite{rahman2018membership}{,} Li et al.\cite{li2021membership} \\ Li and Zhang\cite{li2020label}{,} Rahimian et al.\cite{rahimian2020sampling}{,} Kaya et al.\cite{kaya2020effectiveness}{,} Rezaei and Liu\cite{rezaei2021difficulty} \\ Jia et al.\cite{jia2019memguard}{,} Sablayrolles et al.\cite{sablayrolles2019white}{,} Liu et al.\cite{liu2021ml}{,}  Chang and Shokri\cite{chang2020privacy}\\ Melis et al.\cite{melis2019exploiting}{,} Hui et al.\cite{hui2021practical}{,} Song and Mittal\cite{song2021systematic}{,} Nasr et al.\cite{nasr2019comprehensive}\\ Jayaraman et al.\cite{jayaraman2020revisiting}{,} He et al.\cite{he2021node}{,} Olatunji et al.\cite{olatunji2021membership}{,} Shokri et al.\cite{shokri2021privacy}\\ Chen et al.\cite{chen2020machine}{,} Yaghini et al.\cite{yaghini2019disparate}{,} Choquette et al.\cite{choquette2021label}{,} Leino et al.\cite{leino2020stolen},leaf,text width=19em
            ]
           ]
        ]
        [Generative \\ Models
            [GANs
                [Hayes et al.\cite{hayes2019logan}{,} Hilprecht et al.\cite{hilprecht2019monte}{,} Liu et al.\cite{liu2019performing}{,} Chen et al.\cite{chen2020gan}\\ Wu et al.\cite{wu2019generalization}{,} Mukherjee et al.\cite{mukherjee2021privgan}{,} Webster et al.\cite{webster2021person}
                ,leaf,text width=17em]
            ]
            [VAEs
                 [Hilprecht et al.\cite{hilprecht2019monte}{,} Liu et al.\cite{liu2019performing} Chen et al.\cite{chen2020gan}
                ,leaf,text width=12.5em]
            ]
        ]
        [Regression \\ Models
            [Deep Regression
                [Gupta et al.\cite{gupta2021membership}
                ,leaf,text width=4em]
            ]
        ]
        [Embedding \\ Models
            [NLP Embedding
                [Song and Raghunathan\cite{song2020information}{,} Mahloujifar et al.\cite{mahloujifar2021membership}{,} Thomas et al.\cite{klakow2020investigating}, leaf, text width=17.5em]
            ]
            [Graph Embedding
                [Duddu et al.\cite{duddu2020quantifying},leaf,text width=4em]
            ]
            [Image Encoder
            [Liu et al.\cite{liu2021encodermi},leaf,text width=4em]
            ]
        ]
    ]
    [Adversarial \\ Knowledge\\Level
        [Black-Box
            [Prediction Vector
            [Shokri et al.\cite{shokri2017membership}{,} Yeom et al.\cite{yeom2018privacy}{,} Long et al.\cite{long2018understanding,long2020pragmatic}{,} Yaghini et al.\cite{yaghini2019disparate}\\ Hui et al.\cite{hui2021practical}{,} Song and Mittal\cite{song2021systematic}{,} Jayaraman et al.\cite{jayaraman2020revisiting}{,} Truex et al.~\cite{truex2019demystifying}\\ Liu et al.\cite{liu2019socinf}{,} Chen et al.\cite{chen2020machine}{,} Shokri et al.\cite{shokri2021privacy}{,} Sablayrolles et al.\cite{sablayrolles2019white}
            ,leaf,text width=19em]
            ]
            [Top-K Confidence
            [Shokri et al.\cite{shokri2017membership}{,} Salem et al.\cite{salem2019ml}
            ,leaf,text width=9em]
            ]
            [Label Only
            [Yeom et al.\cite{yeom2018privacy}{,} Li and Zhang\cite{li2020label}{,} Choquette et al.\cite{choquette2021label}{,} Rahimian et al.\cite{rahimian2020sampling},leaf,text width=19.5em]
            ]
        ]
        [White-Box
            [Nasr et al.\cite{nasr2019comprehensive}{,} Melis et al.\cite{melis2019exploiting}{,} Leino et al.\cite{leino2020stolen}{,} Hayes et al.\cite{hayes2019logan}\\ Hilprecht et al.\cite{hilprecht2019monte}{,} Chen et al.\cite{chen2020gan}{,} Rezaei and Liu\cite{rezaei2021difficulty}
            ,leaf,text width=17em]
        ]
    ]
    [Approach \\ Level
        [Classifier \\ Based
            [Shadow Training
                [Shokri et al.\cite{shokri2017membership}{,} Salem et al.\cite{salem2019ml}{,} Long et al.\cite{long2020pragmatic}{,} Shokri et al.\cite{shokri2021privacy}\\ Liu et al.\cite{liu2019socinf}{,} Truex et al.~\cite{truex2019demystifying}{,} Chen et al.\cite{chen2020machine}{,} Chen et al.\cite{chen2020practical}\\ Song and Shmatikov~\cite{song2019auditing}{,} Wang et al.\cite{wang2021edge}
                ,leaf,text width=19em]
            ]
        ]
        [Metric \\ Based
            [Prediction Correctness 
                [Yeom et al.\cite{yeom2018privacy}{,} Choquette et al.\cite{choquette2021label}{,} Irolla and Ch{\^a}tel\cite{irolla2019demystifying}{,} Bentley et al.\cite{bentley2020quantifying}\\Sablayrolles et al.\cite{sablayrolles2019white}{,} ,leaf,text width=19.5em] 
                ]
            [Prediction Loss        
                [Yeom et al.\cite{yeom2018privacy}{,} Sablayrolles et al.\cite{sablayrolles2019white}
                ,leaf,text width=10em]
            ]
            [Prediction Confidence
                [Salem et al.\cite{salem2019ml},leaf,text width=4.5em]
                ]
            [Prediction Entropy
                [Salem et al.\cite{salem2019ml}{,} Song and Mittal\cite{song2021systematic},leaf,text width=10em]
            ]
            [Adversarial Perturbation
                [Li ang Zhang\cite{li2020label}{,} Choquette et al.\cite{choquette2021label}, leaf, text width=10em]
                ]
            [Hypothesis Test
            [Long et al.\cite{long2018understanding}\cite{long2020pragmatic},leaf,text width=5.5em]
            ]    
        ]
        [Differential \\ Comparisons
            [BLINDMI
                [Hui et al.\cite{hui2021practical}
                ,leaf,text width=3.5em]
            ]
        ]
    ]
    [Algorithm \\ Level
        [Centralized
            [Shokri et al.\cite{shokri2017membership}{,} Yeom et al.\cite{yeom2018privacy}{,} Long et al.\cite{long2018understanding,long2020pragmatic}{,} Sablayrolles et al.\cite{sablayrolles2019white}\\ Hui et al.\cite{hui2021practical}{,} Song and Mittal\cite{song2021systematic}{,} Chen et al.\cite{chen2020machine}{,} Jayaraman et al.\cite{jayaraman2020revisiting}\\ Salem et al.\cite{salem2019ml}{,} Li and Zhang\cite{li2020label}{,} Choquette et al.\cite{choquette2021label}{,} Hilprecht et al.\cite{hilprecht2019monte} \\ Hayes et al.\cite{hayes2019logan}{,}  Chen et al.\cite{chen2020gan}{,} Leino et al.\cite{leino2020stolen},leaf,text width=20em]
        ]
        [Federated
            [FedAvg
            [Nasr et al.\cite{nasr2019comprehensive}{,} Lee et al.\cite{lee2021digestive}{,} Zhang et al.\cite{zhang2020gan}{,} Chen et al.\cite{chen2020beyond}{,} Hu et al.\cite{hu2021source},leaf,text width=20em]
            ]
            [FedSGD
            [Melis et al.\cite{melis2019exploiting},leaf,text width=4em]
            ]
        ]
    ]
    [Domain \\ Level
        [Natural \\ Language \\ Processing
            [Text Classification
            [Melis et al.\cite{melis2019exploiting}{,} Liu et al.\cite{liu2019socinf}{,} Wunderlich et al.\cite{wunderlich2021privacy},leaf,text width=14em]
            ]
            [Text Generation
            [Song and Shmatikov\cite{song2019auditing}{,} Hisamoto et al.\cite{hisamoto2020membership},leaf,text width=12em]
            ]
            [Word Embedding
            [Song and Raghunathan\cite{song2020information}{,} Mahloujifar et al.\cite{mahloujifar2021membership}{,} Jagannatha et al.\cite{jagannatha2021membership} \\ Carlini et al.\cite{carlini2020extracting} Thomas et al.\cite{klakow2020investigating} ,leaf,text width=18.5em]
            ]
        ]
        [Computer \\ Vision
            [Image Classification
            [Shokri et al.\cite{shokri2017membership}{,} Yeom et al.\cite{yeom2018privacy}{,} Long et al.\cite{long2018understanding}\cite{long2020pragmatic}{,}  Shokri et al.\cite{shokri2021privacy}\\ Salem et al.\cite{salem2019ml}{,} Truex et al.\cite{truex2019demystifying,truex2019effects} Rahman et al.\cite{rahman2018membership}{,} Leino et al.\cite{leino2020stolen}\\ Li and Zhang\cite{li2020label}{,} Kaya et al.\cite{kaya2020effectiveness}{,} Li et al.\cite{li2021membership}{,}  Liu et al.\cite{liu2021ml}{,} Jia et al.\cite{jia2019memguard}\\ Nasr et al.\cite{nasr2019comprehensive}{,} Choquette et al.\cite{choquette2021label}{,} Chen et al.\cite{chen2020machine}{,} Sablayrolles et al.\cite{sablayrolles2019white}\\ Hui et al.\cite{hui2021practical}{,}  Melis et al.\cite{melis2019exploiting}{,} Song and Mittal\cite{song2021systematic}{,} Rahimian et al.\cite{rahimian2020sampling}\\ Jayaraman et al.\cite{jayaraman2020revisiting}{,}  Rezaei and Liu\cite{rezaei2021difficulty}, leaf,text width=19.5em]
            ]
            [Image Generation
            [Hayes et al.\cite{hayes2019logan}{,} Hilprecht et al.\cite{hilprecht2019monte}{,} Liu et al.\cite{liu2019performing}\\ Chen et al.\cite{chen2020gan}{,} Wu et al.\cite{wu2019generalization}{,}  Mukherjee et al.\cite{mukherjee2021privgan},leaf,text width=13.5em]
            ]
            [Image Segmentation
            [He et al.\cite{he2020segmentations}{,} Shafran et al.\cite{shafran2021reconstruction},leaf,text width=8em]
            ]
        ]
        [Graph
            [Knowledge Graphs 
                [Wang and Sun\cite{wang2021membership},leaf,text width=5em]
            ]
            [Node Classification
                [Duddu et al.\cite{duddu2020quantifying}{,} He et al.\cite{he2021node}{,} Olatunji et al.\cite{olatunji2021membership},leaf, text width=12.5em]
            ]
            [Graph Classification
                [Wu et al.\cite{wu2021adapting},leaf, text width=4.5em]
            ]
        ]
        [Audio
            [Speech Recognition
                [Shah et al.\cite{shah2021evaluating}{,} Miao et al.\cite{miao2021audio},leaf,text width=8em]
            ]
        ]
        [Recommender \\ System
            [Collaborative Filtering
                [Zhang et al.\cite{zhang2021membership},leaf,text width=4em]
            ]
        ]
    ]
  ]
\end{forest}
}
\caption{Taxonomy of membership inference attacks}
\label{taxonomy::attack_category}
\end{figure*}

\begin{table}[hbt!]
\caption{Summary of membership inference attacks work on machine learning models (time ascending).}
\label{table::summary_of_attacks}
\centering
\resizebox{\textwidth}{!}{
\begin{tabular}{ccccccccccc}
\toprule

\textbf{Ref.} & \textbf{Year}& \textbf{Venue} & \textbf{Task} & \textbf{Attack Knowledge} & \textbf{Approach} & \textbf{Baseline} & \textbf{Metric} & \textbf{Dataset}\\
\hline

\cite{shokri2017membership} & 2017 & S\&P & Classification & Black-box & Shadow training & - & \begin{tabular}[c]{@{}c@{}c@{}}  ASR \\ AP, AR \end{tabular}  & \begin{tabular}[c]{@{}c@{}c@{}} Adult \\CIFAR-10, CIFAR-100 \\ Purchase-100, MNIST \\ Texas-100, Foursquare \end{tabular}  \\
\hline

\cite{long2018understanding} & 2018 & arXiv & Classification & Black-box & Hypothesis test & Random guess & AP, AR & Adult, Cancer, MNIST \\
\hline

\cite{yeom2018privacy} & 2018 & CSF & Classification & Black-box & \begin{tabular}[c]{@{}c@{}}  Prediction loss \\ Prediction correctness   \end{tabular} & Shadow training & AP, AR & \begin{tabular}[c]{@{}c@{}}MNIST \\ CIFAR-10, CIFAR-100 \end{tabular}  \\
\hline

\cite{salem2019ml} & 2019 & NDSS & Classification & Black-box & \begin{tabular}[c]{@{}c@{}}  Prediction entropy \\ Prediction confidence  \end{tabular} & Shadow training  & AP, AR & \begin{tabular}[c]{@{}c@{}c@{}}Adult, News \\ MNIST, LFW \\ CIFAR-10, CIFAR-100 \\ Purchase-100, Foursquare  \end{tabular}  \\
\hline

\cite{melis2019exploiting} & 2019 & S\&P & Classification & White-box & Non-zero gradient  &- & AP, AR & \begin{tabular}[c]{@{}c@{}}Yelp-health \\ CSI, Foursquare \end{tabular}  \\
\hline

\cite{nasr2019comprehensive} & 2019 & S\&P & Classification & White-box & \begin{tabular}[c]{@{}c@{}}  Intermediate computation \end{tabular}  & Shadow training &  ASR  & \begin{tabular}[c]{@{}c@{}}CIFAR-100 \\ Texas-100, Purchase-100 \end{tabular}  \\
\hline

\cite{song2019membership} & 2019 & S\&P & Classification & Black-box & \begin{tabular}[c]{@{}c@{}}  Shadow training \\ Prediction confidence  \end{tabular} & - & ASR & \begin{tabular}[c]{@{}c@{}} SVHN \\ CIFAR-10  \end{tabular} \\
\hline

\cite{song2019privacy} & 2019 & CCS & Classification & Black-box & \begin{tabular}[c]{@{}c@{}}  Shadow training \\ Prediction confidence \end{tabular} & - & ASR & \begin{tabular}[c]{@{}c@{}}Fashion-MNIST \\ Yale Face, CIFAR-10  \end{tabular} \\
\hline

\cite{sablayrolles2019white} & 2019 & ICML & Classification & Black-box & Prediction loss & \begin{tabular}[c]{@{}c@{}} Shadow training \\ Prediction correctness  \end{tabular} & ASR & \begin{tabular}[c]{@{}c@{}} CIFAR-10 \\ CIFAR-100  \end{tabular}  \\
\hline

\cite{truex2019demystifying} & 2019 & \begin{tabular}[c]{@{}c@{}}  IEEE Trans. \\ Serv. Comput. \end{tabular} & Classification & Black-box & Shadow training & - &  ASR, AP & \begin{tabular}[c]{@{}c@{}}Adult, CIFAR-10  \\ MNIST, Purchase-100  \end{tabular} \\
\hline

\cite{liu2019socinf} & 2019&  \begin{tabular}[c]{@{}c@{}c@{}}  IEEE Trans.  \\ Comput. \\ Soc. Syst. \end{tabular} &Classification & Black-box & Shadow training & - & ASR, AP & \begin{tabular}[c]{@{}c@{}} Weibo \\ Tweet EmoInt  \end{tabular} \\
\hline

\cite{yaghini2019disparate} & 2019 & arXiv & Classification & Black-box & Subgroup information & - & ASR &  Adult, UTKFace \\
\hline

\cite{hayes2019logan} & 2019 & PoPETs & Generation & \begin{tabular}[c]{@{}c@{}}  Black-box \\ White-box \end{tabular} & Prediction confidence & Random guess & ASR & \begin{tabular}[c]{@{}c@{}}EyePACS \\ CIFAR-10, LFW \end{tabular} \\
\hline

\cite{hilprecht2019monte} & 2019 & PoPETs & Generation & \begin{tabular}[c]{@{}c@{}}  Black-box \\ White-box \end{tabular} & \begin{tabular}[c]{@{}c@{}} Prediction confidence \\ Monte Carlo integration  \end{tabular}  & Logan & ASR, AP & \begin{tabular}[c]{@{}c@{}}Fashion-MNIST \\ MNIST, CIFAR-10 \end{tabular} \\
\hline

\cite{song2019auditing} & 2019 & KDD & Generation & Black-box & Shadow training & - & \begin{tabular}[c]{@{}c@{}}  ASR, AP  \\ AR, AUC \end{tabular}  & \begin{tabular}[c]{@{}c@{}c@{}}SATED, Dislogs \\ Reddit comments \end{tabular} \\
\hline

\cite{liu2019performing} &2019 &ICDM & Generation & White-box & Reconstruction error & - & AUC & \begin{tabular}[c]{@{}c@{}} ChestX-ray8 \\ MNIST, CelebA  \end{tabular} \\
\hline

\cite{chen2020machine} & 2020 & arXiv & Classification & Black-box & Shadow training & - & AUC & \begin{tabular}[c]{@{}c@{}c@{}} Adult\\ Insta-NY, MNIST \\ CIFAR-10, Accident  \end{tabular} \\
\hline

\cite{jayaraman2020revisiting} &2020 & arXiv & Classification & Black-box & Merlin, Morgan & \begin{tabular}[c]{@{}c@{}}  Prediction loss \\ Shadow training \end{tabular} & AP, MA & \begin{tabular}[c]{@{}c@{}} RCV1X, CIFAR-100 \\ Purchase-100, Texas-100  \end{tabular} \\
\hline

\cite{leino2020stolen} & 2020 & \begin{tabular}[c]{@{}c@{}}  USENIX-  \\ Security \end{tabular} & Classification & White-box & Idiosyncratic features & Shadow training & \begin{tabular}[c]{@{}c@{}}  AP  \\ AR, MA \end{tabular}  & \begin{tabular}[c]{@{}c@{}c@{}} Adult \\ Diabetes, LFW \\ Cancer, Hepatitis \\  CIFAR-10, CIFAR-100 \\ MNIST, German credit  \end{tabular} \\
\hline

\cite{hisamoto2020membership} & 2020 & TACL & Generation & Black-box & Shadow training & - & ASR & WMT18 \\
\hline

\cite{he2020segmentations} & 2020 & ECCV & \begin{tabular}[c]{@{}c@{}}  Image  \\ segmentation \end{tabular} & Black-box & Shadow training & Prediction loss & \begin{tabular}[c]{@{}c@{}}  AUC  \\ $\textrm{F}_{1}$-score \end{tabular} & \begin{tabular}[c]{@{}c@{}} Mapillary-Vistas \\ Cityscapes, BDD100K  \end{tabular} \\
\hline

\cite{chen2020gan} & 2020 & CCS & Generation & \begin{tabular}[c]{@{}c@{}}  Black-box \\ White-box \end{tabular} & Reconstruction error & \begin{tabular}[c]{@{}c@{}}  Logan \\ Monte Carlo integration \end{tabular} & AUC & \begin{tabular}[c]{@{}c@{}}CelebA, MIMIC-III\\ Instagram New-York  \end{tabular} \\
\hline

\cite{song2020information} & 2020 & CCS& Embedding & Black-box & Similarity score & Random guess & MA & \begin{tabular}[c]{@{}c@{}} Wikipedia \\ BookCorpus \end{tabular} \\
\hline

\cite{shokri2021privacy} & 2021 & AIES & Classification & Black-box & Model explanations & - & ASR & \begin{tabular}[c]{@{}c@{}c@{}}Adult, Hospital \\ CIFAR-10, CIFAR-100 \\ Purchase-100, Texas-100 \end{tabular}  \\
\hline

\cite{choquette2021label} & 2021 & ICML & Classification & Black-box &\begin{tabular}[c]{@{}c@{}}   Adversarial  \\ perturbation \end{tabular}  & Prediction correctness & ASR, AP & \begin{tabular}[c]{@{}c@{}c@{}} Foursquare \\ Adult, MNIST \\ CIFAR-10, CIFAR-100 \\ Purchase-100, Texas-100 \\  \end{tabular} \\
\hline

\cite{song2021systematic} & 2021 & \begin{tabular}[c]{@{}c@{}}  USENIX-  \\ Security \end{tabular} & Classification & \begin{tabular}[c]{@{}c@{}}  Black-box \\ White-box \end{tabular} & Prediction entropy & Shadow training &\begin{tabular}[c]{@{}c@{}}  ASR  \\ AP, AR \end{tabular} & \begin{tabular}[c]{@{}c@{}}CIFAR-100, Foursquare \\ Purchase-100, Texas-100 \end{tabular} \\
\hline 
 
\cite{hui2021practical} & 2021 & NDSS & Classification & Black-box & \begin{tabular}[c]{@{}c@{}}  Differential \\ comparison \end{tabular} & \begin{tabular}[c]{@{}c@{}c@{}c@{}} Prediction loss \\ Shadow training \\ Prediction confidence \\ Prediction correctness \end{tabular} & $\textrm{F}_{1}$-score & \begin{tabular}[c]{@{}c@{}c@{}}Adult, EyePACS \\ Purchase, Texas-100 \\ CIFAR-100, Bird-200 \\CHMNIST, Foursquare \end{tabular} \\
\hline

\cite{rezaei2021difficulty} & 2021 & CVPR & Classification & White-box & \begin{tabular}[c]{@{}c@{}c@{}c@{}}   Prediction confidence \\ Distance to boundary \\ Intermediate computation \end{tabular} & Prediction correctness & \begin{tabular}[c]{@{}c@{}c@{}}  ASR, AP\\  AR, APR \\ $\textrm{F}_{1}$-score \end{tabular} & \begin{tabular}[c]{@{}c@{}}  MNIST, CIFAR-10 \\ ImageNet, CIFAR-100 \end{tabular} \\
\hline

\cite{duddu2020quantifying} & 2021 & Mobiquitous & Embedding & \begin{tabular}[c]{@{}c@{}}  Black-box \\ White-box \end{tabular} & \begin{tabular}[c]{@{}c@{}}  Shadow training \\ Prediction confidence \end{tabular} & Random guess & ASR, MA & \begin{tabular}[c]{@{}c@{}}  Cora \\ Pubmed, Citesser \end{tabular}  \\
\hline

\cite{hu2021source} & 2021 & ICDM & Classification & White-box & Prediction loss &Random guess & ASR & \begin{tabular}[c]{@{}c@{}}  CIFAR-10 \\ CHMNIST, MNIST \\ Foursquare, Purchase-100 \end{tabular} \\

\bottomrule
\end{tabular}}
\end{table}

\subsection{Taxonomies of Membership Inference Attacks}
To give readers a general picture of MIAs and help readers find the most relevant papers easily, we create a taxonomy of MIAs on ML models in Fig.~\ref{taxonomy::attack_category}. In this taxonomy, we categorize all released papers of MIAs based on different target models, adversarial knowledge, attack approaches, training paradigms, and domains. Specifically, for papers in the category of target model level, we further categorize them based on specific types of the target ML models, i.e., classification models, generative models, regression models, and embedding models. For papers in the category of adversarial knowledge, we further divide them based on whether the MIAs is black-box or white-box attacks. For papers in the category of attack approach level, we further categorize the MIAs into classifier based attacks, metric based attacks, and differential comparisons based attacks. For papers in the category of algorithm level, we further divide them based on whether the target models are trained in a centralized manner or a federated manner. Last, for papers in the category of domain level, we further categorize them based on the specific target domain that the MIAs involve, i.e., natural language processing (NLP), computer vision (CV), graph, audio, and recommemder system. Note that Fig.~\ref{taxonomy::attack_category} not only gives general taxonomies for MIAs according to the above criteria, but also provides detailed characteristics for specific categorized papers. For example, for papers under the category of metric-based attacks, readers can further find the specific metric-based attack approach proposed or involved in the relevant papers, e.g., the paper~\cite{salem2019ml} of Salem et al. proposes the prediction confidence based attack approach.

In addition to categorizing all released papers of MIAs in Fig.~\ref{taxonomy::attack_category}, we further select a few representative papers and list them with their characteristics in Table~\ref{table::summary_of_attacks}. Each of the selected papers either proposes a new MIA or is the first to investigate the membership privacy risks on a specific type of ML model. Compared to Fig.~\ref{taxonomy::attack_category}, Table~\ref{table::summary_of_attacks} gives more information about each paper, which can help readers better understand and compare each paper. Specifically, for each paper listed in Table~\ref{table::summary_of_attacks}, we provide the information of the publication year, the publication venue, the learning task of the target model, the attack knowledge available for the attacker, the specific attack approach, the baseline for the proposed attack, the metrics for evaluating the attack performance, and the datasets used in the experiments. The definition of each metric and a detailed summary of each dataset can be find in Section~\ref{sec06::resources} and Table~\ref{table::dataset_summary} respectively.

\section{Why membership inference attacks work}
\label{sec04::why}

Conducting the theoretical analysis of why membership inference attacks can work is a very challenging task because of the high complexity existing in both training data and target models (especially for deep neural networks). There are some initial works~\cite{yeom2018privacy,farokhi2020modelling,bentley2020quantifying,jha2020extension} formally formulating the problem of MIAs on ML models and providing some theoretical analysis, while the rigorous analysis of why MIAs can work is still in the infant stages. Because most existing literature provides explanations based on practical evaluations, in this section, we discuss why MIAs work from the perspective of the following three aspects mainly based on empirical reasoning.

\noindent \textbf{Overfitting of Target Models. \;} First, many papers~\cite{shokri2017membership,yeom2018privacy,salem2019ml,leino2020stolen,chen2020gan} have pointed out that overfitting of the target ML models is the main factor contributing to the success of MIAs. An ML model is said to overfit to its training data when it performs much better on its training data than test data, i.e., the model cannot generalize well on its test data. The overfitting phenomenon of ML models is usually because of two reasons, i.e., the high model complexity and the limited size of the training dataset~\cite{bishop2006pattern}. Deep learning models such as DNNs are often overparameterized with high complexity, which on the one hand enables them to learn effectively from big data, but on the other hand results in the fact that they may have unnecessarily high capacity of memorizing the noise or the details of a given training dataset~\cite{song2017machine,zhang2021understanding,carlini2019secret,murakonda2020ml}. Moreover, ML models are trained using many (often tens to hundreds) epochs on the same instances repeatedly, rendering the training instances very prone to being memorized by the models. Also, a training dataset with a finite size often fails to represent the whole data distribution, which makes the ML model difficult to generalize to test data, behaving very differently on their training members and non-members. Because MIAs exploit the different behaviors of target ML models on their training data versus test data, ML models overfitted to their training data will be vulnerable to MIAs. Overfitting is sufficient to allow an attacker to perform non-trivial membership inference. For example, if a target classification model is overfitted to its training data that results in a difference (i.e., the generalization gap) between the test accuracy and the train accuracy of the classifier larger than 0, the attacker can easily achieve an overall attack success rate larger than 50\% (i.e., randomly guessing) by leveraging the prediction correctness based attack approach as introduced in Section~\ref{sec::metric-based-attacks}. Note that in~\cite{bentley2020quantifying}, Bentley et al. give a theorem (Theorem~\ref{theorem::overfitting}) that also implies the overfitting of the target models can lead to the performance of an MIA better than randomly guessing (i.e., 50\% attack success rate (ASR)). Theorem~\ref{theorem::overfitting} is as follows, with the symbol denotations and the detailed proof available in the paper~\cite{bentley2020quantifying}:

\begin{theorem}\cite{bentley2020quantifying}
\label{theorem::overfitting}
Given access to a model with generalization gap $g = p_0 - p_1 \ge 0$ (training accuracy minus testing accuracy) and the ratio of training dataset to input domain $q$, there exists a membership inference attack with expected attack success rate (ASR) at least:

$$\begin{array}{l}
ASR \ge \max \{ q,1 - q,q{p_0} + (1 - q)(1 - {p_1})\}, \\
\quad \quad  \ge \max \{ q,1 - q,\min \{ q,1 - q\} (1 + g)\}, \\
\quad \quad  \ge \frac{1}{2}.
\end{array}$$

\end{theorem}

\noindent \textbf{Types of Target Models. \;} Second, the type of target model also plays an important role in the success of MIAs. In general, a target model whose decision boundary is unlikely to be drastically impacted by a particular data record will be more resilient to MIAs. For example, MIAs are evaluated on DNN models, logistic regression models, Naive Bayes models, k-nearest neighbor models, and decision tree models on seven datasets in \cite{truex2019demystifying}. The results show that the decision tree model has the highest attack precision for six datasets, and Naive Bayes models consistently show the lowest precision across all datasets. This is because a single training record only marginally affects the prediction decisions of a given class in Naive Bayes models. By contrast, a record that displays a unique feature can cause a decision tree to grow an entirely new branch, drastically changing the decision boundary. The decision tree models' sensitivity to single records makes MIAs more successful on them. 

\noindent \textbf{Diversity of Training Data. \;} Last, if the training data is more representative, i.e., the training data can better represent the whole data distribution, the target model will be less vulnerable to MIAs. This is because more representative training data can help the target ML model to generalize better on test data. For example, \cite{shokri2017membership} demonstrates a classification model has smaller and smaller attack precision when provided with more and more training records. 

In conclusion, success of MIAs is directly related to three factors: 1) Overfitting of the target model. 2) Type of the target model. 3) Diversity of the target model's training data. Overfitting of the target model is the main reason why MIAs work. Moreover, different ML models remember different amounts of information about their training datasets due to their different structures and training dataset. This leads to different levels of vulnerability to MIAs because different models have different levels of proneness to overfitting.

\section{Membership Inference Defense on Machine Learning Models}
\label{sec05::defense}
In this section, we introduce membership inference defenses on ML models. The existing defenses against MIAs fall into four main categories, i.e., confidence score masking, regularization, knowledge distillation, and differential privacy. 

\subsection{Confidence Score Masking}
Confidence score masking is mainly used to mitigate black-box MIAs on classification models. Confidence score masking aims to hide the true confidence scores returned by the target classifier and thus mitigates the effectiveness of MIAs. There are three methods belonging to this defense category. The first method is that the target classifier does not provide a complete prediction vector but provides top-k confidence scores to the attacker of an MIA. For example, in a classification problem of ten classes, the target classifier only provides the largest three confidence scores when the attacker queries an input record. The second method is that the target classifier only provides the prediction label when the attacker queries an input record. This method provides the most limited knowledge to the attacker. The last method is to add crafted noise to the prediction vector to hide the true confidence scores. The three methods do not need to retrain target classifiers and are only implemented on the prediction vectors, thus they will not influence the target model's accuracy. 

Shokri et al.~\cite{shokri2017membership} evaluate the first two defense methods on a fully connected neural network based classifier on two datasets. They find that restricting the prediction vector to top-$3$ classes does not reduce the attack accuracy of their proposed shadow training based attack. This finding is not surprising because a later paper~\cite{salem2019ml} demonstrates that a black-box binary classifier based attack leveraging partial confidence scores can achieve similar attack performance compared to using complete prediction vectors. Shokri et al.~\cite{shokri2017membership} indeed show that returning only the classifier's predicted label will reduce the attack accuracy. However, as long as the generalization gap exists, a simple prediction correctness based attack will always achieve better attack performance than random guessing. The label-only attacks proposed by Li and Zhang~\cite{li2020label} and Choquette et al.~\cite{choquette2021label} further investigate the membership privacy risks when an attacker gets access only to the target classifiers' predicted labels. Unfortunately, the attacker with only prediction labels can still achieve strong attack performance. Jia et al.~\cite{jia2019memguard} observe that when the attack model is a DNN based binary classifier, it is vulnerable to adversarial examples. Thus, they leverage an adversarial machine learning technique~\cite{kurakin2016adversarial} and propose a defense method called MemGuard. MemGuard adds a carefully crafted noise vector to the prediction vector and turns it into an adversarial example of the attack model. MemGuard does not influence the target models' prediction accuracy while effectively mitigating the black-box DNN based attack to a random guess level. However, Song and Mittal~\cite{song2021systematic} re-evaluate the effectiveness of Memguard using metric based attacks and find that the defended models are still susceptible to membership attacks. 

The advantage of confidence score masking is the simplicity of implementation. It directly works on the trained models' prediction vector and thus does not need to retrain the target model. It is a natural mitigation mechanism against the attacker who uses the complete prediction vector of the target classifier to implement MIAs. However, as we discussed above, confidence score masking might not provide enough privacy guarantees because the label-only attacks still work well, and Memguard is vulnerable to metric based attacks.

\subsection{Regularization}
Regularization aims to reduce the overfitting degree of target models to mitigate MIAs. Therefore, regularization methods that can reduce the overfitting of ML models can be leveraged to defend against MIAs. Existing regularization methods including L2-norm regularization, dropout~\cite{srivastava2014dropout}, data argumentation, model stacking, early stopping, label smoothing~\cite{szegedy2016rethinking}, adversarial regularization~\cite{nasr2018machine}, and Mixup $+$ MMD (Maximum Mean Discrepancy)~\cite{li2021membership} have been proposed and investigated as defense methods in many papers~\cite{shokri2017membership,salem2019ml,song2021systematic,nasr2018machine,li2021membership,hui2021practical,shejwalkar2021membership,kaya2021does,hu2021ear}. Among them, L2-norm regularization, dropout, data argumentation, model stacking, early stopping, label smoothing are classical regularization methods proposed to improve the generalizability of a learned ML model. They are initially proposed to reduce the overfitting of ML models, but they are shown to be quite effective in mitigating MIAs. This is because they help the learned model generalize better to test data and reduce the difference of the model's behaviors on its training data and test data. The adversarial regularization~\cite{nasr2018machine} and Mixup $+$ MMD~\cite{li2021membership} are specially designed regularization techniques that aim to mitigate MIAs. The two proposed methods add new regularization terms to a target classifier's objective function during the training phase and force the classifier to generate similar output distributions for training members and non-members. Adversarial regularization~\cite{nasr2018machine} adds membership inference gain of the attack model as a new regularization term to the objective function of the target model during the training process. The target ML model needs to simultaneously minimize its classification loss and the attack model’s accuracy. The target model is trained in such a way as to preserve its prediction accuracy while mitigating the attacker's performance. Mixup $+$ MMD~\cite{li2021membership} adds the distance between the output distributions of members and non-members computed by Maximum Mean Discrepancy (MMD)~\cite{fortet1953convergence} as a new regularization term to the objective function of the target classifier. The new regularization term forces the classifier to generate similar output distributions for its training members and non-members. As MMD tends to reduce the prediction accuracy of the classifier, the authors in \cite{li2021membership} propose to combine MMD with mix-up training~\cite{zhang2018mixup} to preserve the prediction utility. Note that regularization methods not only work for classification models, and some methods can be used to mitigate MIAs on generation models. For example, Hayes et al.~\cite{hayes2019logan} and Hilprecht et al.~\cite{hilprecht2019monte} demonstrate that dropout can be leveraged as an effective defense method against MIAs on GANs.

Unlike confidence score masking, regularization defends against MIAs no matter whether an attacker is in a black-box or white-box setting. This is because regularization methods change not only the target models' output distribution but also their internal parameters, while methods of confidence score masking only modify models' prediction vectors. Although regularization methods are effective and widely applicable, one drawback of them is that they might not be able to provide satisfactory membership privacy-utility tradeoffs. For example, Shokri et al.~\cite{shokri2017membership} show that L2-norm regularization can mitigate the accuracy of MIAs to random guess level when setting the regularization factor to relatively large values. However, this results in a significant reduction of the target model's prediction accuracy. 

\subsection{Knowledge Distillation}
Knowledge distillation~\cite{ba2014deep,hinton2015distilling} uses the outputs of a large teacher model to train a smaller student model, in order to transfer knowledge from the large model to the small one. It allows the smaller student model to have similar accuracy to their teacher models~\cite{crowley2018moonshine}. Based on knowledge distillation, Shejwalkar and Houmansadr~\cite{shejwalkar2021membership} propose Distillation For Membership Privacy (DMP) defense method. DMP requires a private training dataset and an unlabeled reference dataset. DMP first trains an unprotected teacher model and uses it to label data records in the unlabeled reference dataset. Then, DMP selects data records from the labeled reference dataset that have low prediction entropy to train the target model. The intuition of the selection is that such records are easy to classify and will not be significantly affected by the members of the private training dataset. DMP finally trains a private model based on the selected labeled records. The intuition of DMP is to restrict the private classifier's direct access to the private training dataset, thus significantly reducing the membership information leakage. In contrast to the requirement of a public unlabeled reference dataset in DMP, Zheng et al.~\cite{zheng2021resisting} propose complementary knowledge distillation (CKD) and pseudo complementary knowledge distillation (PCKD) where the transfer data of knowledge distillation all come from the private training set. CKD and PCKD eliminate the need for public data that may be hard to obtain in some applications, making knowledge distillation a more practical defense to mitigate MIAs on ML models.

\subsection{Differential Privacy}
Differential privacy (DP)~\cite{dwork2006calibrating} is a probabilistic privacy mechanism that provides an information-theoretical privacy guarantee. Many papers~\cite{chen2018differentially,yeom2018privacy,rahman2018membership,hayes2019logan,choquette2021label,chen2020gan,leino2020stolen,hui2021practical,shejwalkar2021membership,jia2019memguard,wu2019generalization,jayaraman2019evaluating,truex2019effects,li2021membership,jayaraman2020revisiting,naseri2020toward,ying2020privacy,humphries2020differentially} have applied DP to ML models to mitigate MIAs. When an ML model is trained in a differentially private manner, the learned model does not learn or remember any specific user's details if the privacy budget is sufficiently small. By definition, differenitially private models naturally limit the success probability of MIAs based solely on the model. 

Shokri et al.~\cite{shokri2017membership} first discussed that differentially private models should be able to mitigate MIAs on ML models. Yeom et al.~\cite{yeom2018privacy} theoretically connect DP to MIAs and prove that the membership advantage (refer to Section~\ref{sec06::resources:metrics} Metrics) of an attacker is limited by a function of the privacy budget $\epsilon$. Rahman et al.~\cite{rahman2018membership} first empirically evaluate MIAs on differentially private DNN based classifiers. They find that differentially private models provide privacy protection against strong attackers by only offering poor model utility. Jayaraman and Evans~\cite{jayaraman2019evaluating} further demonstrate that current mechanisms for differentially private ML rarely provide acceptable membership privacy-utility tradeoffs. They comprehensively evaluate MIAs on different variants of the DP mechanisms including differential privacy with advanced composition~\cite{dwork2008differential}, zero concentrated DP~\cite{bun2016concentrated}, and R{\'e}iyi DP~\cite{mironov2017renyi} for ML models. They find that membership privacy leakage is high when setting DP with limited classifiers' accuracy loss, and setting DP to provide strong privacy guarantees, resulting in useless models. Truex et al.~\cite{truex2019effects} evaluate how MIAs differ across classes and how DP affects models when they are trained on skewed data where the class distribution is imbalanced. They report that the minority groups are more vulnerable to MIAs. Moreover, as a mitigation technique, DP tends to decrease a model's utility on the minority groups. Training differentially private ML models is usually achieved by DP-SGD~\cite{abadi2016deep} that adds noise to the gradients of the model during training. Rahimian et al.~\cite{rahimian2020sampling} argue that DP-SGD might significantly hinder the model's prediction performance when the attacker is in the black-box setting. They propose DP-Logits that uses a Gaussian mechanism to only add noise to the logits of the input instance at prediction time and restrict the number of queries. They report that the privacy budget for the DP-Logits is generally lower than the DP-SGD method. 

DP can also be used to defend against MIAs on generative models. Many papers~\cite{chen2018differentially,xu2019ganobfuscator,zhang2018differentially,triastcyn2018generating,beaulieu2019privacy,xie2018differentially,wu2019generalization,nasr2021adversary} have proposed various differentially private generative models to ensure the privacy of their training records. Hayes et al.~\cite{hayes2019logan} first evaluated how MIAs perform on a differentially private GAN proposed by Triastcyn and Faltings~\cite{triastcyn2018generating}. Hayes et al.~\cite{hayes2019logan} find that their proposed white-box MIA achieves great attack performance when $\epsilon$ of the differentially private GAN is relatively high. When $\epsilon$ is small, MIAs perform no better than random guessing. However, small $\epsilon$ also leads GANs to generate bad quality samples. Chen et al.~\cite{chen2018differentially} report similar findings that DP indeed reduces the effectiveness of MIAs on GANs even when $\epsilon$ exceeds practical values (i.e., $\epsilon > 10^{10}$). However, they also mention that DP heavily deteriorates the generation quality of GANs. Moreover, applying DP into training leads to a much higher computation cost where the training time is ten times slower compared to training without DP. Wu et al.~\cite{wu2019generalization} theoretically prove that the generalization gap of GANs trained with differentially private learning algorithms can be bounded. This indicates DP limits the overfitting of GANs to a certain degree and explains why DP helps to mitigate MIAs.

DP provides a theoretical guarantee to protect the membership privacy of training records. DP can be leveraged to mitigate MIAs on both classification models and generative models, no matter whether an attacker is in a black-box or white-box setting. Although DP is widely applicable and effective, one drawback is that it rarely offers acceptable utility-privacy tradeoffs with guarantees for complex learning tasks. That is, it provides meaningless membership privacy guarantees at settings with limited model utility loss, and it results in useless models at settings with strong privacy guarantees~\cite{jayaraman2019evaluating}. However, one must be aware that DP cannot only be used to mitigate MIAs, but also other forms of privacy attacks such as attribute inference attacks~\cite{fredrikson2015model,fredrikson2014privacy} and property inference attacks~\cite{ganju2018property,ateniese2015hacking}. Recent studies have also indicated DP has an interesting connection to model robustness against adversarial examples~\cite{lecuyer2019certified}. 

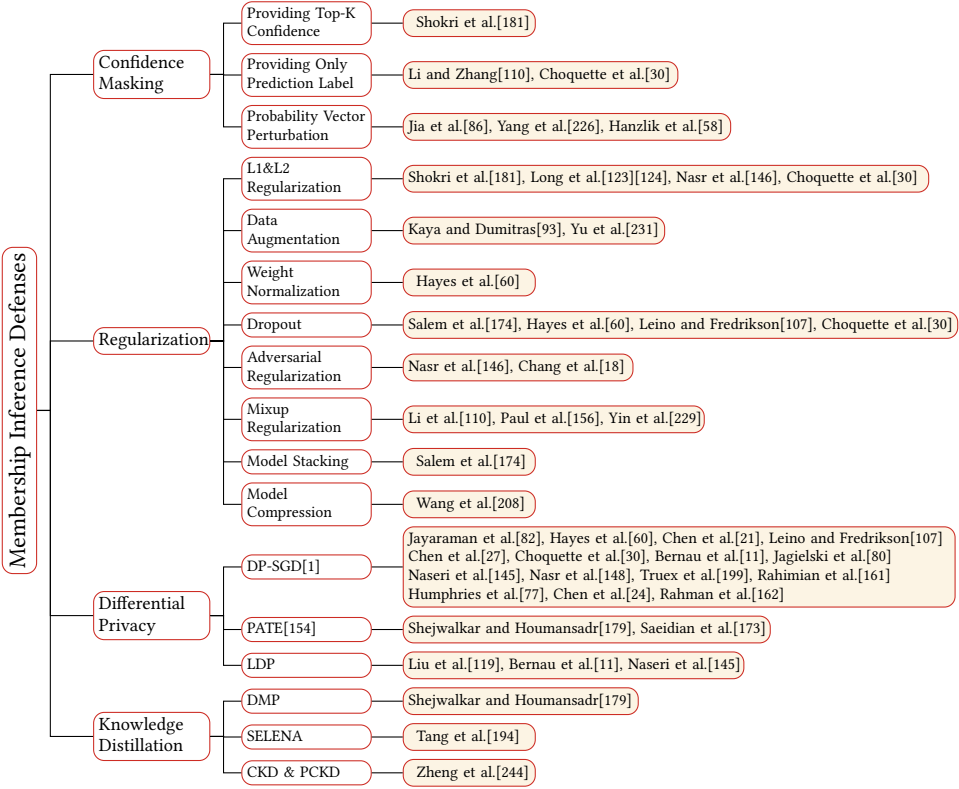
\begin{figure*}[t]
  \centering
\begin{forest}
  forked edges,
  for tree={
  grow=east,
  reversed=true,
  anchor=base west,
  parent anchor=east,
  child anchor=west,
  base=left,
  font=\small,
  rectangle,
  draw=hiddendraw,
  rounded corners,align=left,
  minimum width=2.5em,
s sep=3pt,
inner xsep=2pt,
inner ysep=1pt,
ver/.style={rotate=90, child anchor=north, parent anchor=south, anchor=center},
  },
  where level=1{text width=4em,font=\scriptsize,}{},
  where level=2{text width=4.5em,font=\tiny}{},
  where level=3{text width=3.3em,font=\tiny}{},
  [Membership Inference Defenses, ver
    [Confidence \\ Masking
    [Providing Top-K \\ Confidence
    [Shokri et al.\cite{shokri2017membership},leaf,text width=4em]
    ]
    [Providing Only \\ Prediction Label
    [Li and Zhang\cite{li2021membership}{,} Choquette et al.\cite{choquette2021label},leaf,text width=10em]
    ]
    [Probability Vector \\ Perturbation
    [Jia et al.\cite{jia2019memguard}{,} Yang et al.\cite{yang2020defending}{,} Hanzlik et al.\cite{hanzlik2021mlcapsule},leaf,text width=12em]
    ]
    ]
    [Regularization
    [L1\&L2 \\ Regularization
    [Shokri et al.\cite{shokri2017membership}{,} Long et al.\cite{long2018understanding}\cite{long2020pragmatic}{,} Nasr et al.\cite{nasr2018machine}{,} Choquette et al.\cite{choquette2021label},leaf,text width=19.5em]
    ]
    [Data \\ Augmentation
    [Kaya and Dumitras\cite{kaya2021does}{,} Yu et al.\cite{yu2021does},leaf,text width=9.5em]
    ]
    [Weight \\ Normalization
    [Hayes et al.\cite{hayes2019logan},leaf,text width=4em]
    ]
    [Dropout
    [Salem et al.\cite{salem2019ml}{,} Hayes et al.\cite{hayes2019logan}{,} Leino and Fredrikson\cite{leino2020stolen}{,} Choquette et al.\cite{choquette2021label},leaf,text width=20.6em]
    ]
    [Adversarial \\ Regularization
    [Nasr et al.\cite{nasr2018machine}{,} Chang et al.\cite{chang2019cronus},leaf,text width=8.3em]
    ]
    [Mixup \\ Regularization
    [Li et al.\cite{li2021membership}{,} Paul et al.\cite{paul2021defending}{,} Yin et al.\cite{yin2021defending},leaf, text width=11em]
    ]
    [Model Stacking
    [Salem et al.\cite{salem2019ml},leaf,text width=4em]
    ]
    [Model \\ Compression
    [Wang et al.\cite{wang2021against},leaf,text width=4em]
    ]
    ]
    [Differential \\ Privacy
    [DP-SGD\cite{abadi2016deep}
    [Jayaraman et al.\cite{jayaraman2019evaluating}{,} Hayes et al.\cite{hayes2019logan}{,} Chen et al.\cite{chen2020gan}{,} Leino and Fredrikson\cite{leino2020stolen} \\  Chen et al.\cite{chen2018differentially}{,} Choquette et al.\cite{choquette2021label}{,}  Bernau et al.\cite{bernau2021comparing}{,}  Jagielski et al.\cite{jagielski2020auditting} \\ Naseri et al.\cite{naseri2020toward}{,} Nasr et al.\cite{nasr2021adversary}{,} Truex et al.\cite{truex2019effects}{,} Rahimian et al.\cite{rahimian2020sampling} \\ Humphries et al.\cite{humphries2020differentially}{,} Chen et al.\cite{chen2020differential}{,} Rahman et al.\cite{rahman2018membership},leaf,text width=20.5em]
    ]
    [PATE\cite{papernot2016semi}
    [Shejwalkar and Houmansadr\cite{shejwalkar2021membership}{,} Saeidian et al.\cite{saeidian2021quantifying},leaf,text width=13.5em]
    ]
    [LDP
    [Liu et al.\cite{liu2020secure}{,} Bernau et al.\cite{bernau2021comparing}{,} Naseri et al.\cite{naseri2020toward},leaf,text width=12.5em]
    ]
    ]
    [Knowledge \\ Distillation
    [DMP
    [Shejwalkar and Houmansadr\cite{shejwalkar2021membership},leaf, text width=8.5em]
    ]
    [SELENA
    [Tang et al.\cite{tang2021mitigating},leaf, text width=4em]
    ]
    [CKD \& PCKD
    [Zheng et al.\cite{zheng2021resisting},leaf,text width=4em]
    ]
    ]
  ]
\end{forest}
\caption{Taxonomy of membership inference defenses}
\label{taxonomy::defense_category}
\end{figure*}

\begin{table*}[hbt!]
\caption{Summary of membership inference defenses work on machine learning models (time ascending).}
\label{table::summary_of_defenses}
\centering
\resizebox{\linewidth}{!}{
\begin{tabular}{cccccccccccc}
\toprule

\textbf{Ref.} & \textbf{Year}& \textbf{Venue} & \textbf{Task} & \textbf{Attack Knowledge} & \textbf{Corresp. Attack} & \textbf{Approach} & \textbf{Baseline} & \textbf{Metric} & \textbf{Dataset}\\
\hline

\cite{shokri2017membership} & 2017 & S\&P & Classification & Black-box& Shadow training & \begin{tabular}[c]{@{}c@{}} Top-K confidence \\ L2-regularization \end{tabular} &- & \begin{tabular}[c]{@{}c@{}} ASR \\ AP, AR \end{tabular} & \begin{tabular}[c]{@{}c@{}c@{}c@{}}Purchase-100 \\Adult, Foursquare\\ MNIST, Texas-100  \\ CIFAR-10, CIFAR-100 \\  \end{tabular}  \\
\hline

\cite{nasr2018machine} & 2018 & CCS & Classification & Black-box & Shadow training & \begin{tabular}[c]{@{}c@{}}  Adversarial  \\ regularization \end{tabular}  & L2-regularization & ASR & \begin{tabular}[c]{@{}c@{}c@{}} Texas-100\\ CIFAR-100 \\ Purchase-100  \end{tabular}  \\
\hline

\cite{zhang2018privacy} & 2018 & arXiv & Classification & Black-box & Shadow training & Data obfuscation & Random guess & ASR & CIFAR-10 \\
\hline

\cite{chen2018differentially} & 2018 & arXiv & Classification & Black-box & Shadow training & DP & Random guess & AP & MNIST \\
\hline

\cite{melis2019exploiting} & 2019 & S\&P & Classification & White-box & \begin{tabular}[c]{@{}c@{}}  Non-zero gradient \end{tabular} & \begin{tabular}[c]{@{}c@{}}  Dimensionality  \\ reduction \end{tabular}  & -& AP, AR & \begin{tabular}[c]{@{}c@{}}Yelp-health \\ CSI, Foursquare \end{tabular}  \\
\hline

\cite{salem2019ml} & 2019 & NDSS & Classification & Black-box & \begin{tabular}[c]{@{}c@{}}  Prediction entropy \\ Prediction confidence  \end{tabular} & \begin{tabular}[c]{@{}c@{}}  Dropout  \\ Model stacking \end{tabular}& -  & AP, AR & \begin{tabular}[c]{@{}c@{}c@{}c@{}}Adult, CIFAR-10 \\ News, CIFAR-100 \\ MNIST, Foursquare \\ LFW, Purchase-100  \end{tabular}  \\
\hline

\cite{jayaraman2019evaluating} & 2019 & \begin{tabular}[c]{@{}c@{}}  USENIX-  \\ Security \end{tabular} &  Classification & Black-box & Prediction loss & DP & -& MA &  \begin{tabular}[c]{@{}c@{}}  CIFAR-100 \\ Purchase-100 \end{tabular} \\
\hline

\cite{jia2019memguard} & 2019 & CCS  & Classification & Black-box & Classifier based &  MemGuard  & \begin{tabular}[c]{@{}c@{}c@{}c@{}c@{}} DP \\  Dropout \\  Adversarial \\ regularization \\ Model stacking \\ L2-regularization  \end{tabular} & ASR & \begin{tabular}[c]{@{}c@{}c@{}}   Texas-100 \\ CHMNIST \\ Foursquare \end{tabular} \\
\hline

\cite{hayes2019logan} & 2019 & PoPETs & Generation &  White-box & Prediction confidence & \begin{tabular}[c]{@{}c@{}c@{}}  DP \\Dropout \\ Weight \\ normalization \end{tabular} & - & ASR &  LFW \\
\hline

\cite{wu2019generalization} & 2019 & NeurIPS & Generation & \begin{tabular}[c]{@{}c@{}}  Black-box \\ White-box \end{tabular} & Prediction confidence & DP & - & AUC & LFW, IDC \\
\hline

\cite{yang2020defending} & 2020 & arXiv & Classification & Black-box & \begin{tabular}[c]{@{}c@{}c@{}}  Classifier based \\ Prediction correctness \end{tabular} & \begin{tabular}[c]{@{}c@{}} Prediction \\ purification \end{tabular}  &  \begin{tabular}[c]{@{}c@{}c@{}c@{}}   MemGuard \\  Adversarial \\ regularization \\ Model stacking \end{tabular} & ASR & \begin{tabular}[c]{@{}c@{}c@{}}CIFAR-10 \\ FaceScrub \\ Purchase-100   \end{tabular} \\
\hline

\cite{wu2020characterizing} &2020 & AAAI & Classification & Black-box & Prediction loss & SGLD & Dropout & \begin{tabular}[c]{@{}c@{}c@{}}  ASR \\ AUC  \\ $\textrm{F}_{1}$-score \end{tabular}   & \begin{tabular}[c]{@{}c@{}}  IDC \\ German credit \end{tabular} \\
\hline

\cite{chen2020gan} & 2020 & CCS & Generation & \begin{tabular}[c]{@{}c@{}}  Black-box \\ White-box \end{tabular} & Reconstruction error& DP & - & AUC & \begin{tabular}[c]{@{}c@{}}MIMIC-III\\ CelebA, Insta-NY \\  \end{tabular} \\
\hline

\cite{he2020segmentations} & 2020 & ECCV & \begin{tabular}[c]{@{}c@{}}  Image  \\ segmentation \end{tabular} & Black-box & Shadow training & \begin{tabular}[c]{@{}c@{}c@{}c@{}c@{}}DP \\ Dropout \\ Providing only \\ prediction label \\ Probability vector \\ perturbation \end{tabular} & - & \begin{tabular}[c]{@{}c@{}}  AUC  \\ $\textrm{F}_{1}$-score \end{tabular}  & \begin{tabular}[c]{@{}c@{}c@{}} BDD100K \\ Cityscapes \\ Mapillary-Vistas  \end{tabular} \\
\hline

\cite{tople2020alleviating} & 2020 & ICML & Classification & Black-box & Prediction confidence & Causal learning & Random guess & ASR & \begin{tabular}[c]{@{}c@{}c@{}} Child, Alarm \\ Sachs, Water \\ colored-MNIST  \end{tabular} \\
\hline

\cite{hanzlik2021mlcapsule} & 2021 & CVPR & Classification & Black-box & Prediction entropy & \begin{tabular}[c]{@{}c@{}}Prediction\\ perturbation  \end{tabular}  & - & AUC & CIFAR-100 \\
\hline

\cite{kaya2021does} & 2021 & ICML & Classification & Black-box & prediction loss & \begin{tabular}[c]{@{}c@{}c@{}}Data\\ augmentation \\ Label smoothing  \end{tabular} & -&MA & \begin{tabular}[c]{@{}c@{}c@{}}  CIFAR-10 \\ CIFAR-100 \\ Fashion-MNIST  \end{tabular} \\
\hline

\cite{wang2021against} & 2021 & IJCAI & Classification & Black-box & Shadow training & \begin{tabular}[c]{@{}c@{}}  Model  \\ compression \end{tabular} & \begin{tabular}[c]{@{}c@{}c@{}}  DP  \\ Adversarial \\ regularization \end{tabular} & ASR & \begin{tabular}[c]{@{}c@{}c@{}}  MNIST \\ CIFAR-10 \\ CIFAR-100 \end{tabular} \\
\hline

\cite{shejwalkar2021membership} & 2021 & AAAI & Classification & \begin{tabular}[c]{@{}c@{}}  Black-box  \\ White-box \end{tabular} & \begin{tabular}[c]{@{}c@{}}  Classifier based  \\ Prediction loss \end{tabular} & DMP & \begin{tabular}[c]{@{}c@{}c@{}c@{}c@{}} DP \\ Dropout \\ Adversarial  \\ regularization \\ Weight decay \\ label smoothing \end{tabular} & ASR & \begin{tabular}[c]{@{}c@{}c@{}c@{}} CIFAR-10 \\ Texas-100 \\ CIFAR-100 \\ Purchase-100 \end{tabular} \\
\hline

\cite{mukherjee2021privgan} & 2021 & PoPETs & Generation & White-box & \begin{tabular}[c]{@{}c@{}c@{}}   Monte Carlo \\ integration \\ Prediction confidence \end{tabular} & PrivGAN & - & ASR & \begin{tabular}[c]{@{}c@{}c@{}}   MNIST \\ CIFAR-10 \\Fashion-MNIST \end{tabular} \\
\hline

\cite{chen2021gan} & 2021 & KDD & Generation & \begin{tabular}[c]{@{}c@{}}  Black-box  \\ White-box \end{tabular} & \begin{tabular}[c]{@{}c@{}c@{}}   Monte Carlo \\ integration \\ Prediction confidence \end{tabular} & PAR-GAN & \begin{tabular}[c]{@{}c@{}c@{}}  DP  \\ MIX+GAN \\ PrivGAN \end{tabular} & ASR & \begin{tabular}[c]{@{}c@{}c@{}}   MNIST \\ CIFAR-10 \\ Taxas-100  \end{tabular} \\
\hline

\cite{tang2021mitigating} & 2021 & arXiv & Classification & Black-box & \begin{tabular}[c]{@{}c@{}c@{}}   Classifier based \\ Metric based \end{tabular} &  SELENA & \begin{tabular}[c]{@{}c@{}c@{}}  DP  \\ MemGuard \\ Adversarial \\ regularization  \\ Early stopping  \end{tabular} & ASR, MA & \begin{tabular}[c]{@{}c@{}c@{}}   Texas-100 \\ CIFAR100 \\Purchase-100  \end{tabular} \\

\bottomrule
\end{tabular}}
\end{table*}

\subsection{Taxonomies of Membership Inference Defenses}
Similar to the taxonomies of attacks, we also give readers a general picture of membership inference defenses to help readers find the most relevant papers easily. The taxonomy of membership inference defenses in illustrated in Fig.~\ref{taxonomy::defense_category}. In this taxonomy, we categorize all released papers of membership inference defenses into four main categories, i.e., confidence masking based defenses, regularization based defenses, differential privacy based defenses, and knowledge distillation based defenses. For the papers under each of the categories, we further divide the papers based on the specific defense approach, enabling the readers to find the most relevant papers. In addition to Fig.~\ref{taxonomy::defense_category}, we also select a few representative papers and list them with their characteristics in Table~\ref{table::summary_of_defenses}. Each of the selected papers either proposes a new defense method or comprehensively evaluates the performance of a specific defense method. Compared to Fig.~\ref{taxonomy::defense_category}, Table~\ref{table::summary_of_defenses} gives more information about each paper, which can help readers better understand and compare each paper. Specifically, for each paper in Table~\ref{table::summary_of_defenses}, besides all the characteristics of the papers in Table~\ref{table::summary_of_defenses}, we also introduce the corresponding attacks (i.e., \textbf{Corresp. Attack}) for each paper to help readers be aware of which attacks can be effectively mitigated by the defense proposed in that paper. It is worth noting that some papers appear in both Table~\ref{table::summary_of_attacks} and Table~\ref{table::summary_of_defenses} because they propose an attack approach and a defense approach simultaneously.

\section{Metrics, Datasets, and Open-Source Implementations}
\label{sec06::resources}

\begin{table}[!t]
\centering
\caption{Summary of datasets used for evaluating membership inference attacks and defenses.}
\label{table::dataset_summary}
\resizebox{\linewidth}{!}{%
\begin{tabular}{m{1cm}|llllllm{5.8cm}}
\toprule
\textbf{Type} & \textbf{Task} & \textbf{Dataset} & \textbf{Source} & \textbf{\# Records}  & \textbf{ \# Features } & \textbf{ \# Classes } & \textbf{Paper}\\ 
\hline
\multirow{15}{*}{\begin{tabular}[l]{@{}l@{}}Binary \\ Data \end{tabular}} & \multirow{15}{*}{Classification} 

&  Adult & \cite{Dua:2019}  & 48,842 & 14 & 2 & 
\cite{shokri2017membership,long2018understanding,long2020pragmatic,long2017towards,salem2019ml,truex2019effects,zhang2020privacy,humphries2020differentially,leino2020stolen,choquette2021label,chen2020machine,yaghini2019disparate,farokhi2020modelling,hui2021practical,tonni2020data,rezaei2021accuracy,wang2021edge}
\\
\cline{3-8}

& & Cancer & \cite{Dua:2019} & 699 & 10 & 2 &  \cite{long2018understanding,long2020pragmatic,leino2020stolen} 
\\
\cline{3-8}

& & Diabetes & \cite{Dua:2019} & 768 & 8 & 2 & \cite{leino2020stolen} \\
\cline{3-8}

& & Hepatitis & \cite{Dua:2019} & 155 & 19 & 2 & \cite{leino2020stolen} \\
\cline{3-8}

& & German credit & \cite{Dua:2019} & 1,000 & 20 & 2 & \cite{leino2020stolen,wu2020characterizing} \\
\cline{3-8}

& & Hospital & \cite{strack2014impact} & 101,766 & 127 & 2 & \cite{shokri2021privacy} \\
\cline{3-8}

& & UTKFace & \cite{zhang2017age} & 20,705 & N.A. & 106 & \cite{yaghini2019disparate} \\
\cline{3-8}

& & US-Accident & \cite{moosavi2019accident} & 3,000,000 & 30 & 3 & \cite{chen2020machine} \\
\cline{3-8}

& & Foursquare & \cite{yang2016participatory} & 528,878 & 446 & 30 &

\cite{shokri2017membership,salem2019ml,melis2019exploiting,hui2021practical,jia2019memguard,choquette2021label,song2021systematic,rahimian2020sampling,zhao2021feasibility,zhao2019inferring,duddu2020gecko}
\\
\cline{3-8}

& & Purchase-100 & \cite{purchase} & 197,324 & 600 & 100 & 

\cite{shokri2017membership,long2017towards,salem2019ml,truex2019demystifying,nasr2019comprehensive,song2021systematic,hui2021practical,nasr2018machine,truex2019effects,jayaraman2019evaluating,li2021membership,jayaraman2020revisiting,choquette2021label,bernau2021comparing,shokri2021privacy,jagielski2020auditting,naseri2020toward,nasr2021adversary,rahimian2020sampling,zhao2021feasibility,zhao2019inferring,duddu2020gecko,tonni2020data,shejwalkar2021membership,rezaei2021accuracy,zheng2021resisting,wang2021edge}
\\
\cline{3-8}

& & Texas-100 & \cite{texas} & 67,330 & 6,170 & 100 & 

\cite{shokri2017membership,nasr2019comprehensive,song2021systematic,rahimian2020sampling,hui2021practical,nasr2018machine,jia2019memguard,li2021membership,jayaraman2020revisiting,choquette2021label,bernau2021comparing,shokri2021privacy,tonni2020data,zhang2020privacy,shejwalkar2021membership,rezaei2021accuracy,chen2021gan}
\\
\hline

\multirow{28}{*}{\begin{tabular}[l]{@{}l@{}}Image \\ Data \end{tabular}} &
\multirow{8}{*}{Classification}

& Colored-MNIST & \cite{arjovsky2019invariant} & 70,000 & 28$\times$28$\times$1 & 2 & \cite{tople2020alleviating} \\
\cline{3-8}

& & CH-MNIST &\cite{kather2016multi} & 5,000 & 150$\times$150$\times$1 & 8 & \cite{hui2021practical,jia2019memguard,rahimian2020sampling}
\\
\cline{3-8}

& & SVHN & \cite{svnh} & 99,289 & 32$\times$32$\times$3 & 10 & \cite{song2019membership,jha2020extension,rezaei2021accuracy} \\
\cline{3-8}

& & Yale Face & \cite{lee2005acquiring} & 2,414 & 168$\times$192$\times$1 & 38 & \cite{song2019privacy} \\
\cline{3-8}

& & RCV1X & \cite{lewis2004rcv1} & 800,000 & N.A. & 103 & \cite{jayaraman2020revisiting} \\
\cline{3-8}

& &  Birds-200 &\cite{welinder2010caltech} & 11,788 & N.A. & 200 &\begin{tabular}[l]{@{}l@{}l@{}l@{}} \cite{hui2021practical}
\end{tabular}\\
\cline{3-8}

& & FaceScrub & \cite{ng2014data} & 100,000 & N.A. & 530 & \cite{yang2020defending} \\
\cline{3-8}

& & ImageNet & \cite{deng2009imagenet} & 1,281,167 & N.A. & 1,000 & \cite{sablayrolles2019white,yu2021does,rezaei2021difficulty,rezaei2021accuracy,bagmar2021membership} \\
\cline{2-8}

& \multirow{13}{*}{\begin{tabular}[l]{@{}l@{}}Classification \\ \& Generation \end{tabular}}

& IDC & \cite{janowczyk2016deep} & 277,524 & 50$\times$50$\times$3 & 2 & \cite{wu2019generalization,wu2020characterizing} \\
\cline{3-8}

& & EyePACS & \cite{dr}  & 88,702 & N.A. &5 & 
\cite{hayes2019logan,hui2021practical,paul2021defending}
\\
\cline{3-8}

& & MNIST & \cite{lecun1998gradient}  & 70,000 & 28$\times$28$\times$1 & 10 & 

\cite{shokri2017membership,yeom2018privacy,long2018understanding,long2020pragmatic,salem2019ml,truex2019demystifying,hilprecht2019monte,liu2019performing,li2020label,choquette2021label,rahimian2020sampling,leino2020stolen,rahman2018membership,triastcyn2018generating,truex2019effects,li2021membership,rezaei2021difficulty,chen2018differentially,chen2020machine,hou2019ml,chen2021gan,irolla2019demystifying,nasr2021adversary,zhang2020gan,wang2021against,chen2020beyond,rezaei2021accuracy,mukherjee2021privgan,chen2020practical,wang2021edge}
\\
\cline{3-8}

&  & Fashion-MNIST & \cite{xiao2017fashion}  & 70,000 & 28$\times$28$\times$1 & 10 & 

\cite{song2019privacy,hilprecht2019monte,jagielski2020auditting,irolla2019demystifying,rahimian2020sampling,duddu2020gecko,kaya2020effectiveness,kaya2021does,rezaei2021accuracy,mukherjee2021privgan}
\\
\cline{3-8}

& & CIFAR-10 & \cite{krizhevsky2009learning} &  60,000 & 32$\times$32$\times$3 & 10 & 

\cite{shokri2017membership,yeom2018privacy,salem2019ml,truex2019demystifying,sablayrolles2019white,hayes2019logan,hilprecht2019monte,li2020label,chen2021gan,choquette2021label,rahimian2020sampling,leino2020stolen,rahman2018membership,truex2019effects,li2021membership,kaya2020effectiveness,yin2021defending,rezaei2021difficulty,zhang2018privacy,song2019membership,chen2020machine,shokri2021privacy,jagielski2020auditting,irolla2019demystifying,nasr2021adversary,duddu2020gecko,yu2021does,kaya2021does,wang2021against,jha2020extension,liu2021encodermi,bentley2020quantifying,shejwalkar2021membership,chen2020beyond,rezaei2021accuracy,zheng2021resisting,mukherjee2021privgan,bagmar2021membership,chen2020practical,wang2021edge}
\\
\cline{3-8}

& & CIFAR-100 & \cite{krizhevsky2009learning} &  60,000 & 32$\times$32$\times$3 & 100 & 

\cite{shokri2017membership,yeom2018privacy,salem2019ml,sablayrolles2019white,nasr2019comprehensive,li2020label,choquette2021label,rahimian2020sampling,chen2020practical,leino2020stolen,song2021systematic,hui2021practical,nasr2018machine,jayaraman2019evaluating,li2021membership,kaya2020effectiveness,rezaei2021difficulty,hanzlik2021mlcapsule,shokri2021privacy,naseri2020toward,hou2019ml,truex2019effects,zhao2021feasibility,zou2020privacy,yu2021does,kaya2021does,wang2021against,shejwalkar2021membership,rezaei2021accuracy,zheng2021resisting,bagmar2021membership}\\

\cline{3-8}

& & LFW & \cite{huang2008labeled} &  13,233 & 62$\times$47$\times$3 & 5,749 & 

\cite{salem2019ml,hayes2019logan,melis2019exploiting,li2020label,leino2020stolen,wu2019generalization,mukherjee2021privgan,bernau2021comparing,truex2019effects}
\\
\cline{2-8}

& \multirow{4}{*}{Generation}

& CelebA & \cite{liu2015deep}  & 202,599 & 218$\times$178$\times$3 & 10,177 & 

\cite{liu2019performing,chen2020gan,triastcyn2018generating,liu2021ml,shafran2021reconstruction}
\\
\cline{3-8}

& & MIMIC-III &\cite{johnson2016mimic} & 46,520 & 1,071 & N.A. & \cite{chen2020gan} \\
\cline{3-8}

& & Insta-NY & \cite{backes2017walk2friends} & 34,336 & 4,048& N.A. & \cite{chen2020gan,chen2020machine} \\
\cline{3-8}

& & ChestX-ray8 & \cite{wang2017chestx}  & 108,948 & 1024$\times$1024$\times$1 &32,717 & 

\cite{liu2019performing}
\\
\cline{2-8}

& \multirow{3}{*}{Segmentation}

& Cityscapes & \cite{cordts2016cityscapes} & 20,000 & N.A. & 30 & \cite{he2020segmentations,shafran2021reconstruction} \\
\cline{3-8}

& & BDD100K & \cite{yu2018bdd100k} & 100,000 & N.A. & N.A. & \cite{he2020segmentations} \\
\cline{3-8}

& & Mapillary-Vistas & \cite{neuhold2017mapillary} & 25,000 & N.A. & 37 & \cite{he2020segmentations} \\
\hline

\multirow{12}{*}{\begin{tabular}[l]{@{}l@{}}Text \\ Data \end{tabular}} 
& \multirow{6}{*}{Classification} 

&  CSI & \cite{verhoeven2014clips}  & 1,412 & N.A. & 2 & 
\cite{melis2019exploiting}
\\
\cline{3-8} 

& & Review & \cite{mcauley2013hidden} & 364,038 & N.A. & 2 & \cite{chen2020comprehensive,yin2021defending} \\
\cline{3-8}

& & Tweet EmoInt & \cite{mohmmad2017wassa} & 7,097 & N.A. & 4 & \cite{liu2019socinf} \\
\cline{3-8}

& &  Yelp-health & \cite{melis2019exploiting}  & 17,938 & N.A. & 10 & 

\cite{melis2019exploiting}
\\
\cline{3-8} 

& &  News & \cite{lang1995newsweeder}  & 20,000 & N.A. & 20 & 

\cite{salem2019ml}
\\
\cline{3-8} 

& & Weibo & \cite{weibo} & 23,000 & N.A. & N.A. & \cite{liu2019socinf} \\
\cline{2-8}

& \multirow{4}{*}{Generation}  

& Reddit comments & \cite{reddit} & 83,293 & N.A. & N.A. & \cite{song2019auditing} \\
\cline{3-8}

& & Dialogs & \cite{danescu2011chameleons} & 220,579 & N.A. & N.A. & \cite{song2019auditing} \\
\cline{3-8}

& &SATED & \cite{michel2018extreme} & 2,324 & N.A. & N.A. & \cite{song2019auditing} \\
\cline{3-8}

& & WMT18 & \cite{wmt} & N.A. & N.A. & N.A. & \cite{hisamoto2020membership} \\
\cline{2-8}

& \multirow{2}{*}{Embedding}

& Wikipedia & \cite{mahoney2011large} & 150,000 & N.A. & N.A. & \cite{song2020information} \\
\cline{3-8}

& & BookCorpus & \cite{zhu2015aligning} & 14,000 & N.A. & N.A. & \cite{song2020information} \\
\hline

\multirow{4}{*}{\begin{tabular}[l]{m{8cm}} Graph \\ Data \end{tabular}} 
& \multirow{4}{*}{Classification}

& Pubmed &\cite{sen2008collective} &19,717 & 500 & 3 & \cite{duddu2020quantifying,olatunji2021membership} \\
\cline{3-8}

& &  Citeseer&\cite{sen2008collective} & 3,327 & 3,703 & 6 & \cite{duddu2020quantifying,he2021node,olatunji2021membership} \\
\cline{3-8}

& & Cora&\cite{sen2008collective} & 2,708 & 1,433 & 7 & \cite{duddu2020quantifying,he2021node,olatunji2021membership} \\
\cline{3-8}

& & Lastfm & \cite{rozemberczki2020characteristic} & 7,624 & 7,842 & 18 & \cite{he2021node} \\

\bottomrule
\end{tabular}
}
\end{table}


In this section, we first summarize the metrics for evaluating attack and defense performance of membership inference. Then, we summarize common datasets used in membership inference attack and defense works on machine learning models. Lastly, we provide links to the open-source implementation of popular methods.
\subsection{Metrics}
\label{sec06::resources:metrics}
In this subsection, we first briefly introduce the general evaluation metrics of target models. Then, we give a detailed introduction of particular evaluation metrics designed for attacks and defenses. 
\subsubsection{General Metric}
According to Table~\ref{table::summary_of_attacks} and Table~\ref{table::summary_of_defenses}, many existing works tackle the binary or multi-class classification problem. The metric \textbf{Accuracy} for classification problems is used by existing works to reflect the classification performance of target models. Readers can refer to \cite{wikipedia_matrix} for a detailed explanation of \textbf{Accuracy}. Another metric \textbf{Generalization Error}~\cite{hardt2016train} defined as absolute difference between the Train Accuracy and the Test Accuracy of the target model is used by existing works to reflect the overfitting level of target models. A larger Generalization Error indicates the target model is more overfitted to its training data, demonstrating the target model is associated with higher privacy risks of membership inference attacks~\cite{shokri2017membership}.

\subsubsection{Adversarial Metric}
Besides the general metrics used for evaluating target models, a number of metrics which measure the attack and defense performance have been proposed or used by existing works. In this subsection, we introduce the detailed formulations and descriptions of widely used metrics. Each metric name follows the reference of the first paper that proposes or uses this metric, and the references inside the parentheses refer to other attack and defense papers using this metric.
\begin{itemize}

\item \textbf{Attack Success Rate (ASR)~\cite{shokri2017membership}} (\cite{nasr2019comprehensive,salem2019ml,hayes2019logan,nasr2018machine,jia2019memguard,song2019privacy,sablayrolles2019white,song2019auditing,truex2019demystifying,zhang2018privacy,liu2019socinf,song2019membership,hilprecht2019monte,choquette2021label,bernau2021comparing,shokri2021privacy,yaghini2019disparate,naseri2020toward,song2021systematic,hisamoto2020membership,yang2020defending,tople2020alleviating,hou2019ml,irolla2019demystifying,truex2019effects,chang2020privacy,zhang2020gan,duddu2020quantifying,duddu2020gecko,kaya2020effectiveness,wu2020characterizing,zou2020privacy,yu2021does,he2021node,tonni2020data,miao2021audio,wang2021against,chen2020differential,rezaei2021difficulty,jha2020extension,bentley2020quantifying,shejwalkar2021membership,grosse2021killing,lee2021digestive,chen2020beyond,zheng2021resisting,webster2021generating,bagmar2021membership,chen2020practical,wang2021edge,chen2021gan,yin2021defending,liu2021encodermi,tang2021mitigating} ). ASR is the most commonly used metric to measure the performance of a given attack approach:
\begin{equation*}
    \textrm{ASR} = \frac{\textrm{\# Successful attacks}}{\textrm{\# All attacks}}.
\end{equation*}

\item \textbf{Attack Precision (AP)~\cite{shokri2017membership}} (\cite{yeom2018privacy,salem2019ml,melis2019exploiting,long2018understanding,long2020pragmatic,song2019auditing,truex2019demystifying,leino2020stolen,liu2019socinf,song2019membership,hilprecht2019monte,choquette2021label,song2021systematic,jayaraman2020revisiting,chen2020comprehensive,zhang2020gan,zou2020privacy,miao2021audio,rezaei2021difficulty,bentley2020quantifying,zheng2021resisting,olatunji2021membership,chen2020practical,liu2021encodermi}). AP is the fraction of records classified as members that are indeed members of the training dataset:
\begin{equation*}
    \textrm{AP} = \frac{\textrm{\# Members correctly classified as members}}{\textrm{\# Records classified as members}}.
\end{equation*}


\item \textbf{Attack Recall (AR)~\cite{shokri2017membership}} (\cite{yeom2018privacy,salem2019ml,melis2019exploiting,long2018understanding,song2019auditing,leino2020stolen,song2019membership,song2021systematic,nasr2021adversary,zou2020privacy,miao2021audio,rezaei2021difficulty,bentley2020quantifying,zheng2021resisting,olatunji2021membership,chen2020practical,liu2021encodermi}).
AR is the fraction of the training dataset’s members that are correctly classified as members:
\begin{equation*}
    \textrm{AR} = \frac{\textrm{\# Members correctly classified as members}}{\textrm{\# All members}}.
\end{equation*}

\item \textbf{Attack False Positive Rate (FPR)~\cite{rezaei2021difficulty}}(\cite{nasr2021adversary}).
Attack FPR is the fraction of the testing dataset’s records that are misclassified as members:
\begin{equation*}
    \textrm{FPR} = \frac{\textrm{\# Non-members classified as members}}{\textrm{\# All non-members}}.
\end{equation*}

\item \textbf{Membership Advantage (MA)~\cite{yeom2018privacy}} (\cite{jayaraman2019evaluating,leino2020stolen,song2020information,jayaraman2020revisiting,farokhi2020modelling,li2021membership,duddu2020quantifying,kaya2020effectiveness,kaya2021does,humphries2020differentially,wunderlich2021privacy,tang2021mitigating}).
MA is the difference between the Attack Recall (AR) and the attack False Positive Rate (FPR):
\begin{equation*}
    \textrm{MA} = \textrm{AR} - \textrm{FPR}.
\end{equation*}

\item \textbf{Attack $\textrm{F}_{1}$-score~\cite{rahman2018membership}} (\cite{he2020segmentations,zhang2018privacy,hui2021practical,zhang2020gan,wu2020characterizing,miao2021audio,rezaei2021difficulty}) Attack $\textrm{F}_{1}$-score is the harmonic mean of Attack Precision and Attack Recall:
\begin{equation*}
    \textrm{F}_1\textrm{-score} = \frac{2 \cdot \textrm{AP} \cdot \textrm{AR} }{\textrm{AP} +\textrm{AR} }.
\end{equation*}

\item \textbf{Attack AUC~\cite{song2019auditing}}(\cite{chen2020gan,hanzlik2021mlcapsule,wu2019generalization,he2020segmentations,liu2019performing,chen2020comprehensive,rahimian2020sampling,zhao2021feasibility,zhao2019inferring,wu2020characterizing,zou2020privacy,shafran2021reconstruction,wunderlich2021privacy,rezaei2021accuracy,olatunji2021membership}). AUC is {Area-under-the-ROC-curve}. Readers can refer to \cite{wikipedia_auc} for a detailed explanation of AUC. Attack AUC is sensitive to the probability rank of members, which is larger when members are ranked higher than non-members according to the predicted probability of a membership inference binary classifier.

\end{itemize}

\subsection{Evaluation Datasets and Open-Source Implementations}
Table~\ref{table::dataset_summary} summarizes all datasets used in membership inference attack and defense works on ML models. We categorize all datasets based on different data types. Among the binary datasets, Adult, Foursquare, Purchase-100, and Texas-100 have been widely used as classification benchmarks in many papers. Among the image datasets, MNIST, Fashion-MNIST, CIFAR-10, CIFAR-100, and LFW are widely used as both classification and generation benchmarks. Because MIAs are relatively unexplored on NLP, different works use different text datasets to evaluate membership privacy risks on different models. Among graph datasets, Citeseer and Cora are widely used node classification benchmarks. Due to the limited space of this paper, we provide links to the open-source implementation of popular methods in our GitHub repository. We hope our work can facilitate the community to move towards the construction of benchmarks, similar to other areas~\cite{deng2009imagenet,wang2018glue}.

\section{Future Directions}
\label{sec07::future_directions}

In this section, we discuss several main challenges and potential research opportunities on membership inference attacks and defenses in order to inspire interested readers to explore this field more.

\subsection{Membership Inference Attacks}

\begin{enumerate}
    \item The assumption that the target ML models are heavily overfitted to their training data both lacks the practicability of MIAs and limits their applicability, while it underpins the success of most existing works of MIAs. The attacker cannot guarantee that the target models are always heavily overfitted because many regularization techniques has been used to prevent the overfitting of the ML models. Since MIAs on non-overfitted ML models are pretty challenging, they have not been explored in depth and thus inspire a practically interesting direction for MIAs. The feasibility of MIAs on such non-overfitted models still remains unknown in existing literature. It is even really difficult for an attacker to tell if a target ML model is overfitted or not as she has quite limited knowledge about the training process.

    \item Recently, self-supervised learning models like Bert~\cite{devlin2018bert}, T5~\cite{raffel2020exploring}, and MoCo~\cite{he2020momentum} are popular and have achieved promising results for many complex downstream tasks such as computer vision and natural language processing, but MIAs on such emerging models have not been explored yet. It is urgent and crucial to investigate the membership privacy risks on self-supervised learning models because their training datasets consist of large unlabeled data such as image, text, and audio without labelling, which can still be highly private and unauthorized. It can be intriguing to adopt the principles and designs in existing MIAs on supervised and unsupervised learning schemes for the increasingly important self-supervised learning models.

    \item Adversarial machine learning aims to fool or misguide a model with malicious input with typical examples like data poisoning attacks and model evasion attacks, and plays an increasingly important role in applications such as auto-driving safety and spam filtering. While adversarial ML and MIAs as two separate research areas have developed in parallel, it is interesting and challenging to understand their relationships in terms of their theoretical foundations, algorithmic designs, building blocks, etc. For instance, how to explain the phenomenon that the behaviour difference between non-members and members in MIAs is similar with that between adversarial and benign examples in adversarial ML will bridge the two areas to achieve private and secure ML. There have been a couple of initial works~\cite{choquette2021label,li2020label} exploiting the fact that the training data is more robust against adversarial attacks than the test data to launch label-only MIAs in the black-box context. One specific research question can be how the attacker creates MIAs by leveraging the techniques of white-box adversarial examples, e.g., Fast Gradient Sign Method (FGSM)~\cite{goodfellow2014explaining} and Projected Gradient Descent (PGD)~\cite{madry2017towards}.
    
    \item There are more avenues where MIAs have not been explored but intensive research efforts are demanded due to their high importance, e.g., contrastive learning models and meta-learning models. Contrastive learning aims to learn similar/dissimilar representations from data that are organized into similar/dissimilar pairs~\cite{chen2020improved}. Meta-learning, also known as learning to learn, refers to the process of improving a learning algorithm over multiple learning episodes~\cite{hospedales2020meta}. The particular training paradigms of contrastive learning and meta-learning are pretty different from the conventional supervised learning scheme and therefore impose unique challenges on MIAs. For example, data augmentation is a core building block for contrastive learning for generalizable embedding features by enriching positive training examples with perturbation. An interesting question naturally arises: How does the data augmentation in contrastive learning affect MIAs? In meta-learning, the training dataset consists of a certain number of source tasks, and each source task has both training data and validation data. How do MIAs behave on the data of different source tasks? Moreover, for each source task, do MIAs behave differently on the training data and validation data?
    
    \item As discussed in Section~\ref{sec::federated-learning}, federated learning has emerged as a promising privacy-aware paradigm, and some initial works~\cite{melis2019exploiting,nasr2019comprehensive,lee2021digestive,zhang2020gan} have demonstrated the feasibility of MIAs on federated learning. However, the applicability of existing MIAs is limited to homogeneous federated learning where each local party is assumed to have the same model architecture, while such an assumption is too strong for real applications because the computation and communication capabilities of each party can vary significantly and dynamically~\cite{li2020federated}. On the contrary, heterogeneous federated learning schemes such as FedMD~\cite{li2019fedmd} and HeteroFL~\cite{diao2020heterofl} have recently been proposed to handle the system heterogeneity~\cite{li2020federated} without requiring local models to share the same architecture. While heterogeneous federated learning is more practically realistic, little research has been done in depth for exploring the membership privacy risks in this learning paradigm. Thus, extensive efforts are required to investigate the feasibility and efficacy of MIAs on heterogeneous federated learning schemes, so as to shed light on building more private federated learning systems. Another interesting question is how we can exploit the system heterogeneity information to perform inference attacks on a specific party, since a recent pioneer work~\cite{hu2021source} has shown the success of source inference attacks in the homogeneous context.

    \item It is of practical interest to investigate new applications by exploiting the information gained from MIAs, which we believe is still in its infancy. The following are a few recently emerging examples we listed to inspire more applications. Based on the membership information from MIAs, the recently proposed source inference attacks in federated learning can further identify  the party (i.e., the source) owning a given training member~\cite{hu2021source}. Because training data are more prone to evasion attacks in the context of adversarial learning, the membership information from MIAs can also be leveraged to improve the design of adversarial examples. Another interesting application of membership inference is to audit if a data record has contributed to the training of an ML model. This is an essential step for data owners to achieve the full control of their data as described by many recently issued  privacy regulations and laws such as GDPR~\cite{wikipedia_gdpr}, because unauthorised use of their data for training an ML model can be detected with membership inference on the trained model. Further actions like machine unlearning~\cite{bourtoule2021machine} can be taken if the data owners wish to recall the contribution of their data to the trained model.
    
\end{enumerate}

\subsection{Membership Inference Defenses}

\begin{enumerate}

    \item As revealed in existing literature, the overfitting of ML models is the main factor contributing to the success of MIAs and the level of overfitting can be leveraged to measure the effectiveness of a membership inference defense method, but it is still a challenge to capture the overfitting phenomenon, especially for unsupervised learning. Most existing effective defenses can mitigate MIAs because they are designed to reduce the overfitting level of the target ML models. With labelled validation data, we can estimate generalization error~\cite{mohri2018foundations} to monitor the overfitting level for the supervised learning models and can further determine if a defense method is effective or not. However, it is still a challenge for unsupervised learning models to perceive and handle overfitting due to the lack of data labels, and this considerably limits the potential of designing and evaluating the defenses against MIAs on such models. For instance, no defense has been proposed to mitigate MIAs on word embedding models, to the best of our knowledge. Thus, it is a promising research direction to explore the defenses on unsupervised learning models from the perspective of overfitting, given that the unsupervised learning models like GANs~\cite{goodfellow2014generative} and VAEs~\cite{kingma2013auto} have become increasingly popular and important in many applications where abundant unlabelled data are available.

    \item With the rapid development of generative models such as GANs and VAEs, the generated examples from these models can be highly similar to the original training data, and it is very intriguing to explore the possibility of leveraging the generated examples as surrogate data for model training so as to mitigate MIAs. The surrogate datasets can help decouple the direct relationship between the original training data and the target model output, while the population features can still be retained to train effective models. Besides, data augmentation techniques can also be used to create surrogate datasets by perturbing original training examples. Extensive efforts are required to explore the theoretical foundation for this category of membership inference defenses and more defense mechanisms are expected based on a wide range of generative models and data augmentation techniques. It is worth noting that many emerging machine learning schemes can not only be the targets of MIAs, but they can also be exploited to defend against MIAs.
    
    \item While utility is the ultimate goal in many data analytic applications, it is very challenging to design defense solutions with an acceptable trade-off between membership privacy and model utility. The defense methods offering strong privacy guarantees often come at the cost of high utility loss of the target model. In particular, the deployed query interface of the target models has been ignored in most existing works, while this factor plays an important role in the utility-privacy trade-off as it determines if a white-box or black-box attack can be launched. Taking the classification model as an example, existing differential privacy based defenses often add a large amount of noise to the gradients of the target classifier in a white-box manner during the training process, and thus can heavily lower its prediction accuracy. However, if the classification model is deployed by only providing the query interface to users in a black-box manner, can we achieve little prediction performance sacrifice while adding noise only to the output of the target classifier while guaranteeing the differential privacy requirements?

    \item Even though federated learning emerges as a privacy-aware learning paradigm, it faces an increasing number of privacy attacks~\cite{lyu2020threats}, and it is urgent for the community to develop the corresponding defense techniques. Differential privacy is well known to offer strong privacy guarantees and can be integrated in federated learning, but the model utility can be retained only when the number of parties is very large~\cite{mcmahan2017learning,geyer2017differentially}. It is still a challenge to achieve acceptable privacy-utility trade-off when applying differential privacy to business-to-business federated learning~\cite{yang2019federated} where the number of parties is usually small. Based on MIAs, source inference attacks~\cite{hu2021source} have been specifically designed against federated learning, but no specific defense techniques have been proposed, which offers many appealing opportunities for interested researchers in this field.
\end{enumerate}

\section{Conclusion}
\label{sec08::conclusion}
In this work, we have covered most, if not all, the released papers about membership inference attack and defense on ML models. We first give the definition of MIAs on ML models and introduce existing attack approaches. We give a taxonomy to categorize all the papers of MIAs. Next, we discuss why MIAs can work on ML models. Then, we introduce the existing defense approaches used to mitigate MIAs and give a taxonomy to categorize the papers of membership inference defense. We have summarized most existing evaluation metrics, datasets, and open-source implementations of popular approaches. Finally, for both membership inference attack and defense, we discuss the challenges and point out the potential research opportunities for future studies. Through this comprehensive survey, we hope to prepare a solid foundation for future research in this field.

\bibliographystyle{ACM-Reference-Format}
\bibliography{reference.bib}

\end{document}